\documentclass[11pt]{article}
\usepackage[preprint]{acl}
\usepackage{times}
\usepackage{latexsym}
\usepackage[T1]{fontenc}
\usepackage[utf8]{inputenc}
\DeclareUnicodeCharacter{266B}{\textmusicalnote}
\usepackage{microtype}
\IfFileExists{inconsolata.sty}{\usepackage{inconsolata}}{}
\usepackage{graphicx}
\usepackage{amsmath,amsthm,mathtools}
\usepackage{newtxmath}
\usepackage{algorithm}
\usepackage[noend]{algpseudocode}
\usepackage{booktabs}
\usepackage{multirow}
\usepackage{xcolor}
\usepackage{colortbl}
\newtheorem{lemma}{Lemma}

\definecolor{evopink}{HTML}{D6337F}
\newsavebox{\objtabbox}
\newcommand{\sd}[1]{\,{\scriptsize$\pm$#1}}


\newcommand{\Rorch}{R\textsuperscript{2}Orch}
\title{\Rorch{}: Recursive Self-Improvement for Agent Orchestration\\ via Self-Play within a Skill-Internalization Loop}
\author{Mingda Zhang$^{1}$\quad Yun Huang$^{1}$\quad Qiang Huang$^{2}$\quad Yanjin Li$^{3}$\quad Rui Qian$^{2}$\\
\bfseries Qichao Zhao$^{4}$\quad Tiesunlong Shen$^{5}$\quad Na Zhao$^{1}$\quad Jian Wang$^{1}$\quad Xiaoying Tang$^{1}$\\[4pt]
\mdseries\footnotesize $^{1}$The Chinese University of Hong Kong, Shenzhen, China\quad $^{2}$Fudan University, China\quad $^{3}$University of Illinois Urbana-Champaign, USA\\
\mdseries\footnotesize $^{4}$Tsinghua University, China\quad $^{5}$National University of Singapore, Singapore\\[3pt]
\mdseries\footnotesize\texttt{mingdazhang@ieee.org}\quad\texttt{qiianruii@gmail.com}}
\begin{document}
\maketitle
\begin{abstract}

In recent years, agentic orchestration has been widely applied to solve complex human tasks.
However, existing orchestration learning still faces key challenges, including \emph{hand-designed curricula}, \emph{advantage collapse} when a task's rollouts all reach the same outcome, and \emph{unreliable evaluation signals} that judge each skill by one noisy score.
To address these challenges, we propose \Rorch{}, a new paradigm of \emph{Recursive Self-Play Orchestration}: an inner loop of self-play that improves the orchestrator on self-chosen tasks, and an outer loop that improves how the inner loop improves.
To support this paradigm, we introduce \emph{Recursive Learnability}, a training objective that rewards a Proposer for choosing tasks whose rollouts keep reaching different outcomes as skills are withdrawn.
Built on this objective, we further propose \emph{Skill Withdrawal}, in which the outer loop withdraws each skill once a controlled evaluation within self-play shows that the orchestrator performs as well without it.
Moreover, the two loops form one cycle: failures are distilled into skills, skills into parameters, and withdrawals lead the Proposer to new failures, so the orchestrator moves toward zero-skill deployment.
Experimental results on twelve datasets show that \Rorch{} consistently outperforms baselines across question answering, math, code, and interactive tasks.
Our code is available at \href{https://github.com/beita6969/R2Orch}{{\urlstyle{same}\nolinkurl{github.com/beita6969/R2Orch}}}.

\end{abstract}
\section{Introduction}
\begin{figure}[t]
\centering
\includegraphics[width=\columnwidth]{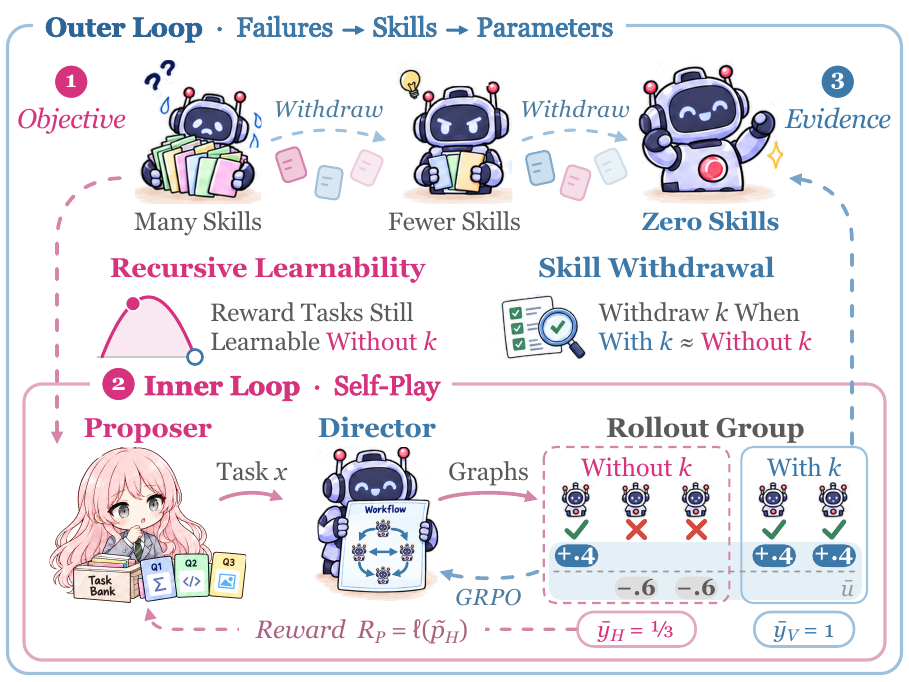}
\caption{A recursion needs two channels: the outer loop sets the inner loop's objective, and the inner loop, Proposer--Director self-play on a randomized rollout group, returns the evidence for withdrawing skills.}
\label{fig:intro}
\end{figure}
\begin{figure*}[t]
\centering
\includegraphics[width=\textwidth]{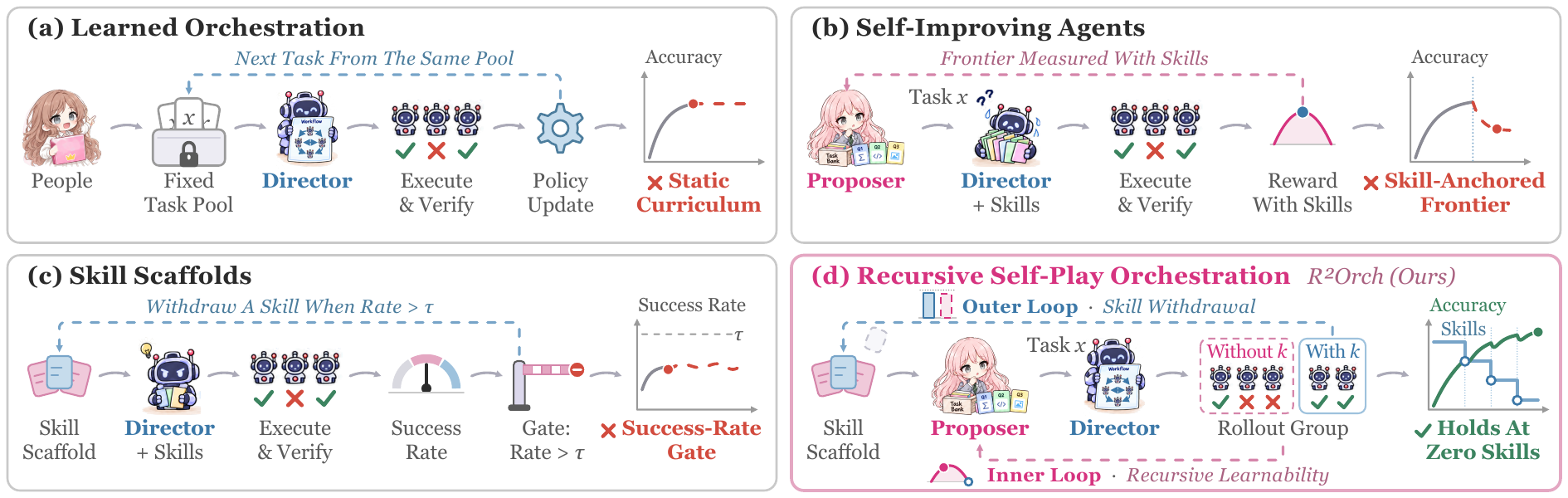}
\caption{Paradigm comparison. (a) Learned orchestration trains on a hand-designed \emph{static curriculum}. (b) Self-improving agents measure the frontier with skills present, a \emph{skill-anchored frontier} where advantages collapse once skills leave. (c) Skill scaffolds withdraw guidance through \emph{success-rate gates}, unreliable evaluation signals pinned by self-play. (d) \Rorch{} nests two loops joined by downward (objective) and upward (evidence) channels.}
\label{fig:paradigms}
\end{figure*}

\suppressfloats[t]

In recent years, a variety of powerful agentic systems have been applied to solve a wide range of human problems \citep{zhang2025aflow,li2026flow,zhang2026flowsteer}, gradually moving beyond hand-written agent pipelines toward orchestrators that learn to compose agent teams.

In this process, the agent graph has become a key bridge from task goals to executable collaboration: by giving each node a role and each edge a hand-off, orchestrators solve tasks with improved controllability, compositionality, and reusability \citep{zhang2025aflow,zhang2026flowsteer}, as shown in Figure~\ref{fig:intro}.
However, in practice, orchestrators still learn from tasks people collect and lean on in-prompt skills \citep{li2026flow,zhang2026skillflow}, making them costly to scale and hard to improve on their own.

To address these issues, three main lines of work have emerged, as shown in Figure~\ref{fig:paradigms}.
First, \emph{learned orchestration} optimizes workflows and planning from task feedback \citep{zhang2025aflow,li2026flow,zhang2026flowsteer}, taking the task pool as given.
Second, \emph{self-improving agents} refine their own memory, harness code, answers, or tasks through repeated loops \citep{zhu2026rsiagent,srikanth2026recursive,zhang2026hyperagents,lu2026arex,fu2026self}.
Third, \emph{skill scaffolds} hand procedural knowledge to a policy \citep{zhang2026skillflow,shi2026pats,lu2026skill0,shen2026dynamic}, in context at inference or as training-time support.

However, these methods still face three challenges (Figure~\ref{fig:paradigms}a--c).
\textbf{(i) Hand-designed curricula.} Learned orchestration and skill internalization practice on task pools that people collect in advance \citep{zhang2026flowsteer,shi2026pats}, and self-play chooses its own tasks while the solver keeps retrieving its skills \citep{fu2026self}, so choosing tasks and withdrawing skills do not inform each other.
\textbf{(ii) Advantage collapse.} Group-based RL such as GRPO learns from differences among a task's rollouts, which vanish when they all reach the same outcome \citep{he2026advantage}; self-play keeps tasks at the solver's frontier with skills in context \citep{fu2026self}, so withdrawing a skill pushes the tasks it supported toward all-wrong rollouts, and advantage collapse returns.
\textbf{(iii) Unreliable evaluation signals.} Training-time scaffolds fade guidance on a success rate mixing task choice with the skill's effect \citep{shi2026pats}, and skill libraries keep or retire skills on one noisy score each, so they drift until injecting a skill scores worse than injecting nothing \citep{zhang2026library,zhang2605ratchet}.

To address these challenges, we propose \textbf{\Rorch{}} (Fig.~\ref{fig:paradigms}d), a new paradigm of \emph{Recursive Self-Play Orchestration}, in which an inner loop of self-play improves the orchestrator, a Director, on tasks a Proposer chooses, and an outer loop improves how the inner loop improves.
Two channels join the loops, one carrying the objective down and one the evidence up.
To support this paradigm, we introduce \emph{Recursive Learnability}, the downward channel, which rewards the Proposer for choosing tasks whose rollouts keep reaching different outcomes as skills are withdrawn, scoring each task with the candidate skill withheld.
Built on this objective, we further propose \emph{Skill Withdrawal}, the upward channel, in which a random part of each rollout group hides one skill, turning self-play into a controlled evaluation whose Bayesian posterior over the skill's remaining benefit decides when it can leave.
The two loops thus close into one cycle: failures are distilled into skills, skills into parameters, and withdrawals lead the Proposer to new failures, so the Director improves on its own and moves toward zero-skill deployment.

We evaluate \Rorch{} on twelve benchmarks in question answering, math, code, and interactive tasks.
Results show that \Rorch{} outperforms all baselines on every benchmark, improves six other Worker backbones without retraining, and keeps its accuracy when tested with zero skills.
Together, these results mark a step toward the recursive self-improvement of agent orchestrators.

\section{Related Work}

\paragraph{Learned Orchestration and Self-Play.}
Prior work searches agent workflows \citep{zhang2025aflow}, learns graph editing \citep{zhang2026flowsteer}, or trains the planner of an agent system \citep{li2026flow}, over tasks that people collect in advance.
Self-play lets a Proposer pose tasks to a solver, rewarding regret \citep{dennis2020emergent}, solver uncertainty \citep{huang2026r}, or learning progress while distilling failures into skills \citep{fu2026self}.
Group-based RL \citep{shao2024deepseekmath,liu2025understanding} gives no advantage to a task whose rollouts all reach the same outcome \citep{he2026advantage}, so curricula favor learnable tasks \citep{rutherford2024no} or filter the rest out \citep{yu2026dapo}.
\Rorch{} scores tasks with the candidate skill withheld, so its Proposer picks tasks whose rollouts keep reaching different outcomes instead of filtering collapsed ones.

\paragraph{Skill Evolution and Self-Improvement.}
Skills enter the prompt at inference \citep{zhang2026skillflow} or support training and are withdrawn on a success-rate threshold \citep{shi2026pats}, a preset budget \citep{lu2026skill0}, or leave-one-skill-out validation \citep{shen2026dynamic}.
Rollouts with and without a skill score it \citep{tu2026dynamic,lin2026skillc}, paired tests accept self-modifications \citep{shawn2026pace}, and one noisy score per skill lets a library drift \citep{zhang2605ratchet,zhang2026library}.
Recursive self-improvement systems revise their own memory, code, or answers \citep{zhu2026rsiagent,srikanth2026recursive,lu2026arex,zhang2026hyperagents,wang2026metaskill} or adopt a self-modification once it is shown to help \citep{schmidhuber2007godel,zhang2026darwin}.
\Rorch{} nests self-play in a skill-withdrawal loop and decides each withdrawal from the controlled evaluation within every randomized rollout group, at no extra rollouts.

\begin{figure*}[t]
\centering
\includegraphics[width=\textwidth]{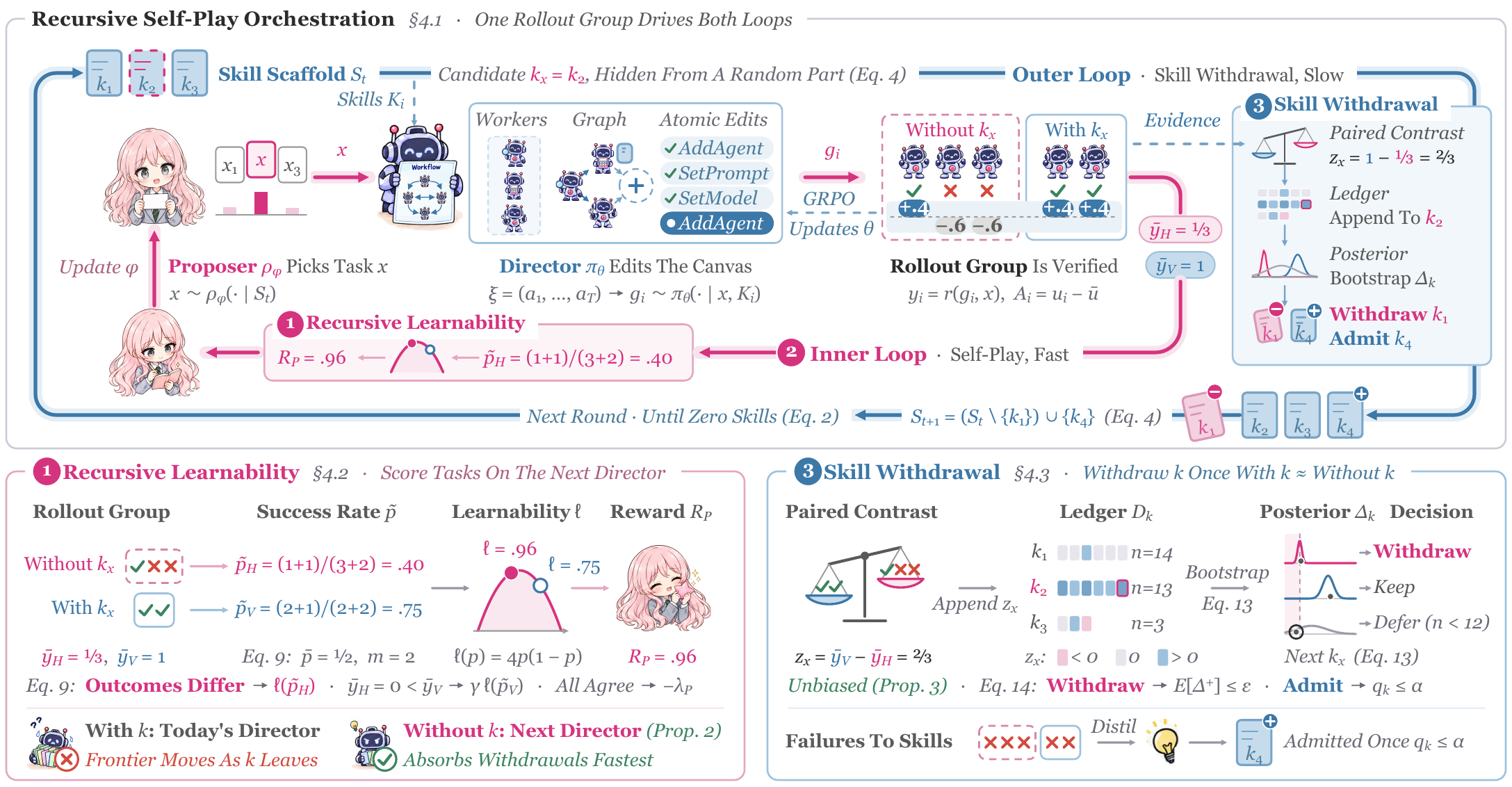}
\caption{Overview of \Rorch{}. Top: Proposer--Director self-play (inner loop) and skill withdrawal (outer loop) share one randomized rollout group. Bottom left: Recursive Learnability scores a task with the candidate skill hidden. Bottom right: Skill Withdrawal withdraws or admits each skill by the posterior of its paired contrast.}
\label{fig:method}
\end{figure*}

\section{Preliminaries}

\noindent\textbf{Definition 1: Agent Graph.}
An agent graph is a directed graph $\mathcal G=(V,E,\kappa,o)$, where each node $v\in V$ is an agent with a role prompt and a model route, each edge $e\in E$ is a directed or bidirectional hand-off between two agents, $\kappa$ assigns every agent a layer, and $o\in V$ is the output agent whose answer is returned (our role-prompted nodes extend the operator nodes of \citealp{zhang2026flowsteer}).

\noindent\textbf{Definition 2: Director Trajectory.}
A complete sequence of $T$ atomic edits from an empty agent graph to \textsc{Finish} defines a trajectory, in which every edit is validated and applied before the next one and its feedback returns to the Director:
\begin{equation}
\begin{gathered}\xi=(a_1,\dots,a_T)\Rightarrow g,\\ \mathcal G_t=\textsc{Apply}(\mathcal G_{t-1},a_t),\ \ \mathcal G_0=\varnothing,\ \ g=\mathcal G_T,\end{gathered}
\label{eq:traj}
\end{equation}
where $a_t$ is one of \textsc{AddAgent}, \textsc{SetPrompt}, \textsc{SetModel}, \textsc{SetLayer}, \textsc{ConsiderRelation}, \textsc{DeleteAgent}, \textsc{SetOutput}, and \textsc{Finish}, and $g$ is the final agent graph that frozen Workers execute.

\noindent\textbf{Problem Statement.}
Given an evaluation distribution $\rho$, a frozen Worker pool, a verifier $r$, and a skill library, we seek a Director $\pi_\theta$ that builds a successful agent graph with no skill in its context, trained on tasks chosen by a Proposer $\rho_\phi$:
\begin{equation}
\max_{\theta}\ \mathbb{E}_{x\sim\rho}\,\mathbb{E}_{g\sim\pi_\theta(\cdot\mid x,\varnothing)}\big[r(g,x)\big],
\label{eq:target}
\end{equation}
where $r(g,x)\in\{0,1\}$ is the verifier's correctness after the Workers execute $g$ on $x$, and the empty set marks zero-skill deployment.
Skills scaffold training, and the Proposer chooses tasks whose rollouts reach different outcomes.
Equation~\eqref{eq:target} is the deployed target; Section~\ref{sec:method} shows how a shrinking scaffold and a learned curriculum approach it.

\section{Method}
\label{sec:method}

As illustrated in Figure~\ref{fig:method}, this section introduces the \Rorch{} framework, including Recursive Self-Play Orchestration (Section~\ref{sec:paradigm}), Recursive Learnability (Section~\ref{sec:rl}), and Skill Withdrawal (Section~\ref{sec:sw}), all built on one randomized rollout group.

\subsection{Recursive Self-Play Orchestration}
\label{sec:paradigm}

Recursive Self-Play Orchestration nests two self-improvement loops (Figure~\ref{fig:method}, top).
In the inner loop, self-play improves the Director on tasks that the Proposer chooses; the outer loop improves how the inner loop improves, by withdrawing the Director's skills as each stops helping.

\noindent\textbf{Two time scales.}
Following two-time-scale stochastic approximation \citep{borkar1997stochastic}, both policies take proximal best responses on the fast time scale, each a clipped step \citep{schulman2017proximal}:
\begin{equation}
\begin{gathered}
\theta_{t+1}\approx\arg\max_{\theta}\mathbb E\,v_K(x;\theta)-\tfrac1\beta\,\mathrm{KL}(\pi_\theta\|\pi_{\theta_t}),\\
\phi_{t+1}\approx\arg\max_{\phi}\mathbb E\,R_P(x)-\tau\,\mathrm{KL}(\rho_\phi\|\rho_0).
\end{gathered}
\label{eq:nested}
\end{equation}
Here the Director, trained on the Proposer's picks with $K\sim q_x$, is scored by the value $v_K(x;\theta)=\mathbb E_{\xi\sim\pi_\theta(\cdot\mid x,K)}[u(\xi,x)]$ of the context it reads under the reward $u$ (correctness less a Worker cost), and the Proposer by the reward $R_P$ of Section~\ref{sec:rl}.
On the slow scale, the outer loop updates the scaffold from the same rollout groups, and the scaffold in turn sets the randomized context of every group:
\begin{equation}
\begin{gathered}
\mathcal S_{t+1}=\Phi\big(\mathcal S_t;\,\{\Pi_k\}_{k\in\mathcal S_t},\,\mathcal F_t\big),\\
q_x=\tfrac{n_h}{G}\,\delta_{K_x\setminus k_x}+\tfrac{n_v}{G}\,\delta_{K_x},
\end{gathered}
\label{eq:outer}
\end{equation}
where $K_x\subseteq\mathcal S_t$ are the skills retrieved for $x$, $k_x\in K_x$ is the skill under test, $\Pi_k$ the posterior over $k$'s benefit, $\mathcal F_t$ the groups in which both parts fail, and $\Phi$ the decisions of Section~\ref{sec:sw}; of the $G=n_h+n_v$ rollouts, rollout $i$ reads $K_i$, which is $K_x\setminus k_x$ in the hidden part $\mathcal H_x$ ($n_h>n_v$) and $K_x$ in the visible part $\mathcal V_x$.
Because $\mathcal S_t$ constrains both $q_x$ and the context on which $R_P$ is computed, each outer step rewrites the objectives of the inner loop.

\noindent\textbf{Director update.}
The randomized context is also how the inner loop prepares each withdrawal: because $q_x$ puts most of its mass on $K_x\setminus k_x$, most of each update already trains the Director without the candidate skill.
The Director uses the group-relative estimator of GRPO \citep{shao2024deepseekmath,liu2025understanding} with this adaptation; with centered advantages, it is a leave-one-out gradient \citep{ahmadian2024back} of the first objective in Eq.~\eqref{eq:nested}:
\begin{equation}
\begin{aligned}
&\nabla_\theta\,\mathbb E_{K\sim q_x}\big[v_K(x;\theta)\big]=\mathbb E\Big[\tfrac{1}{G-1}\textstyle\sum_{i=1}^{G}\big(u_i-\bar u\big)\\
&\qquad\cdot\nabla_\theta\log\pi_\theta\big(\xi_i\mid x,\,K_i\big)\Big],
\end{aligned}
\label{eq:grpo}
\end{equation}
where $\xi_i$ is the edit trajectory of rollout $i$ (Definition~2) and $u_i-\bar u=(1-\tfrac1G)(u_i-\bar u_{-i})$ is the leave-one-out baseline of rollout $i$.

\noindent\textbf{What each loop contributes.}
To see how the two loops share the work, let $J_{\mathcal S}(\theta)=\mathbb E_{x\sim\rho}[p_{K_x}(x)]$ be the success rate under scaffold $\mathcal S$ on $\rho$, with $p_K(x)$ the success probability under context $K$. Over the rounds it telescopes into inner gains $\Gamma_t=J_{\mathcal S_t}(\theta_{t+1})-J_{\mathcal S_t}(\theta_t)$ and outer costs $C_t=J_{\mathcal S_t}(\theta_{t+1})-J_{\mathcal S_{t+1}}(\theta_{t+1})$:
\begin{equation}
J_{\mathcal S_T}(\theta_T)=J_{\mathcal S_0}(\theta_0)+\textstyle\sum_{t<T}\big(\Gamma_t-C_t\big).
\label{eq:decomp}
\end{equation}
Here $\Gamma_t$ is what self-play adds on a fixed scaffold, and $C_t$ is what the scaffold change takes away: for one withdrawn skill $k$, its benefit $p_{K_x}-p_{K_x\setminus k}$ weighted by how often $k$ is retrieved.
With $\mathcal S_T=\varnothing$ the left side is the zero-skill objective, and the other two components each serve one term: Recursive Learnability drives $\Gamma_t$ (Section~\ref{sec:rl}), and Skill Withdrawal limits the withdrawal cost in each $C_t$ (Section~\ref{sec:sw}).

\noindent\textbf{Proposition 1.} \emph{Zero-skill success is the starting success plus inner gains less outer costs, and self-play tracks each scaffold's equilibrium.}
\emph{Proof.} Appendix~\ref{app:proofs}, via Lemmas~\ref{lem:decomp} and~\ref{lem:twotime}, following the two-time-scale analysis of \citet{borkar1997stochastic}.

\subsection{Recursive Learnability}
\label{sec:rl}

Recursive Learnability is the downward channel, through which the outer loop sets the inner loop's objective: it rewards the Proposer for tasks whose rollouts keep reaching different outcomes with the candidate skill withheld (Figure~\ref{fig:method}, bottom left).

\noindent\textbf{Tilt and its log-partition.}
How much a task can raise its success follows from the proximal step itself.
Following relative-entropy policy search \citep{peters2010relative} and KL-regularized RL as inference \citep{korbak2022rl}, a step of size $\beta$ tilts the Director toward success, and its log-partition is the outcome's cumulant generating function:
\begin{equation}
\begin{gathered}
\pi_\beta(g\mid x,K)=\pi_\theta(g\mid x,K)\,e^{\beta r(g,x)}/Z_K,\\
\log Z_K=\beta p_K+\tfrac{\beta^2}{2}p_K(1-p_K)+O(\beta^3).
\end{gathered}
\label{eq:tilt}
\end{equation}
Here $Z_K(x;\beta)=1+p_K(e^{\beta}-1)$ with $p_K=p_K(x)$, so the slope and curvature of $\log Z_K$ at $\beta=0$ are the success probability and the outcome variance.
A step gains $\beta\,p_K(1-p_K)$ in success at a KL cost of $\beta^2p_K(1-p_K)/2$ to leading order, so within a KL budget the gain grows with the curvature.

\noindent\textbf{Recursive target.}
On the current Director, this curvature is the learnability of the task \citep{rutherford2024no}.
The progress that counts, however, is what survives withdrawal (Eqs.~\eqref{eq:grpo} and~\eqref{eq:decomp}), so Recursive Learnability takes the curvature on the context with the candidate skill withheld:
\begin{equation}
\begin{gathered}
\ell_{\mathrm{rec}}(x)\coloneqq4\,\partial^2_\beta\log Z_{K_x\setminus k_x}\big|_{\beta=0}\\
=4\,p^{-}\big(1-p^{-}\big)=\tfrac{4}{G-1}\,\mathbb E\big[\textstyle\sum_{i}(y^{-}_i-\bar y^{-})^2\big],
\end{gathered}
\label{eq:recl}
\end{equation}
a learnability proportional to the expected advantage energy of the group, where $p^{-}=p_{K_x\setminus k_x}(x)$, $y^{-}_i$ are the outcomes of $G$ rollouts that read $K_x\setminus k_x$ with mean $\bar y^{-}$, and $\ell(p)=4p(1-p)$.
The target vanishes exactly when $p^{-}\in\{0,1\}$, where every group collapses, and peaks at one half; since each withdrawal removes one more skill from $K_x\setminus k_x$, it follows the scaffold down to the zero-skill Director.

\noindent\textbf{Proposer reward.}
The hidden part of each group already reads $K_x\setminus k_x$, so it measures $p^{-}$ at no extra cost.
The reward is an empirical-Bayes plug-in estimate of $\ell_{\mathrm{rec}}$ on this part \citep{gelman1995bayesian}, discounted by $\gamma<1$ when only the visible part progresses and penalized by $\lambda_P$ on collapse:
\begin{equation}
\begin{gathered}
R_P(x)=(1-c_x)\,\gamma^{h_x}\,\ell(\tilde p_x)-\lambda_P\,c_x,\\
\tilde p_x=\mathbb E\big[p\mid y_{\mathcal A_x}\big]=\frac{\sum_{i\in\mathcal A_x}y_i+m\,\bar p}{|\mathcal A_x|+m}.
\end{gathered}
\label{eq:rp}
\end{equation}
Here $c_x=\mathbf1[\bar y\in\{0,1\}]$ flags collapse, $h_x=\mathbf1[\bar y_{\mathcal H}=0<\bar y_{\mathcal V}]$ switches $\mathcal A_x$ from $\mathcal H_x$ to $\mathcal V_x$, and the Beta prior shrinks toward the dataset's running hidden-part success rate $\bar p$ with strength $m$.

\noindent\textbf{Proposer update.}
These rewards become a curriculum through the Proposer.
As a KL-regularized environment designer \citep{dennis2020emergent}, it maximizes $J_P(\rho)=\mathbb E_{x\sim\rho}[R_P(x)]-\tau\,\mathrm{KL}(\rho\|\rho_0)$ over the candidate list $\mathcal C_t$, whose optimum is the closed-form Gibbs curriculum (Lemma~\ref{lem:proposer}):
\begin{equation}
\rho^{\star}(x)=\frac{\rho_0(x)\,e^{\bar R_P(x)/\tau}}{\sum_{x'\in\mathcal C_t}\rho_0(x')\,e^{\bar R_P(x')/\tau}}.
\label{eq:gibbs}
\end{equation}
With $\bar R_P(x)$ the expected reward of task $x$, the curriculum tilts the base model $\rho_0$ toward tasks of high Recursive Learnability.
A pairwise update, with each pick as the other's baseline \citep{ahmadian2024back}, moves the Proposer toward these tasks.

\noindent\textbf{Proposition 2.} \emph{Under a small trust region, a task's success gain grows with its Recursive Learnability.}
\emph{Proof.} Appendix~\ref{app:proofs}, via Lemmas~\ref{lem:tilt}--\ref{lem:energy} on tilt and energy (graph-determined outcomes).

\begin{table*}[t]
\centering
\renewcommand{\arraystretch}{1.04}\setlength{\tabcolsep}{3pt}
\resizebox{\textwidth}{!}{
\begin{tabular}{@{}ll@{\hspace{5pt}}ccccc@{\hspace{5pt}}c@{\hspace{5pt}}c@{\hspace{5pt}}c@{\hspace{5pt}}ccc@{}}
\toprule
\multicolumn{2}{c}{} & \multicolumn{5}{c}{\textbf{Question answering}} & \textbf{Math} & \textbf{Code} & \textbf{Interactive} & \multicolumn{3}{c}{\textbf{Average}}\\
\cmidrule(lr){3-7}\cmidrule(lr){8-8}\cmidrule(lr){9-9}\cmidrule(lr){10-10}\cmidrule(lr){11-13}
\multicolumn{2}{l}{\textit{(a) In-distribution}} & \multicolumn{2}{c}{\textbf{HotpotQA}} & \multicolumn{2}{c}{\textbf{NQ-Open}} & \textbf{MedQA} & \textbf{AIME 2026} & \textbf{MBPP+} & \textbf{ALFWorld} & \multicolumn{3}{c}{\textbf{IID}} \\
\cmidrule(lr){3-4}\cmidrule(lr){5-6}\cmidrule(lr){7-7}\cmidrule(lr){8-8}\cmidrule(lr){9-9}\cmidrule(lr){10-10}\cmidrule(lr){11-13}
\textbf{Type} & \textbf{Method} & \footnotesize EM & \footnotesize F1 & \footnotesize EM & \footnotesize F1 & \footnotesize Acc. & \footnotesize Acc. & \footnotesize Pass@1 & \footnotesize SR & \footnotesize EM & \footnotesize F1 & \footnotesize Acc. \\
\midrule
\multirow{2}{*}{\small\textit{Backbone}} & Qwen3.5-9B & 57.81\sd{0.00} & 73.75\sd{0.22} & 23.59\sd{0.86} & 32.89\sd{1.39} & 70.94\sd{0.86} & 48.67\sd{2.98} & 78.59\sd{0.70} & 47.19\sd{0.70} & 40.70\sd{0.43} & 53.32\sd{0.70} & 61.35\sd{0.84} \\
 & DeepSeek-V4-Flash & 71.25\sd{1.28} & 82.50\sd{0.85} & 38.44\sd{0.86} & 49.77\sd{0.89} & 84.38\sd{0.00} & 49.33\sd{3.65} & 84.69\sd{0.70} & 60.16\sd{1.10} & 54.84\sd{0.71} & 66.14\sd{0.62} & 69.64\sd{1.07} \\
\cmidrule(lr){1-13}
\multirow{2}{*}{\small\textit{Training}} & SFT & 63.28\sd{1.10} & 77.58\sd{0.89} & 25.78\sd{0.00} & 37.50\sd{0.19} & 73.59\sd{0.86} & 34.67\sd{1.83} & 78.28\sd{0.86} & 40.94\sd{0.43} & 44.53\sd{0.55} & 57.54\sd{0.46} & 56.87\sd{0.59} \\
 & GRPO & 69.22\sd{0.43} & 81.33\sd{0.43} & 24.53\sd{0.70} & 32.27\sd{1.41} & 77.03\sd{0.70} & 34.00\sd{4.94} & 82.19\sd{0.65} & 50.94\sd{1.02} & 46.88\sd{0.48} & 56.80\sd{0.74} & 61.04\sd{1.60} \\
\cmidrule(lr){1-13}
\multirow{1}{*}{\small\textit{Search}} & AFlow & 87.66\sd{0.65} & 88.83\sd{0.67} & 76.56\sd{0.78} & 81.09\sd{0.63} & 89.22\sd{0.35} & 52.00\sd{2.98} & 85.31\sd{0.86} & 73.44\sd{0.00} & 82.11\sd{0.33} & 84.96\sd{0.46} & 74.99\sd{0.92} \\
\cmidrule(lr){1-13}
\multirow{2}{*}{\small\textit{Agent RL}} & AgentFlow & 86.25\sd{0.89} & 88.44\sd{1.98} & 81.41\sd{0.86} & 85.78\sd{0.98} & 88.44\sd{1.69} & 61.33\sd{1.83} & 85.63\sd{0.43} & 76.09\sd{0.70} & 83.83\sd{0.59} & 87.11\sd{1.10} & 77.87\sd{0.35} \\
 & FlowSteer & 87.97\sd{0.70} & 88.91\sd{1.49} & 83.91\sd{0.43} & 86.88\sd{0.49} & 90.94\sd{0.70} & 67.33\sd{3.65} & 88.91\sd{1.02} & 82.19\sd{0.86} & 85.94\sd{0.28} & 87.90\sd{0.78} & 82.34\sd{1.00} \\
\cmidrule(lr){1-13}
\multirow{2}{*}{\small\textit{Skill int.}} & SKILL0 & 87.34\sd{0.86} & 88.52\sd{1.49} & 82.97\sd{0.35} & 86.48\sd{0.87} & 89.53\sd{0.70} & 64.67\sd{6.91} & 87.97\sd{1.18} & 83.44\sd{0.35} & 85.16\sd{0.39} & 87.50\sd{0.86} & 81.40\sd{1.65} \\
 & PATS & 89.06\sd{1.46} & 91.25\sd{1.59} & 84.22\sd{1.16} & 87.19\sd{1.21} & 91.41\sd{1.46} & 74.00\sd{3.65} & 90.63\sd{0.55} & 87.97\sd{0.70} & 86.64\sd{0.80} & 89.22\sd{1.00} & 86.00\sd{0.95} \\
\cmidrule(lr){1-13}
\multirow{2}{*}{\small\textit{Self-play}} & SESA & 88.13\sd{1.40} & 90.39\sd{1.23} & 83.28\sd{0.43} & 86.95\sd{0.67} & 88.75\sd{0.70} & 68.00\sd{5.06} & 88.44\sd{0.86} & 81.09\sd{0.65} & 85.70\sd{0.59} & 88.67\sd{0.70} & 81.57\sd{1.31} \\
 & MetaSkill & 87.03\sd{1.42} & 88.75\sd{0.64} & 82.50\sd{0.70} & 85.63\sd{0.53} & 88.44\sd{0.86} & 55.33\sd{2.98} & 86.25\sd{1.52} & 79.06\sd{0.86} & 84.77\sd{0.83} & 87.19\sd{0.42} & 77.27\sd{1.21} \\
\midrule
\rowcolor{evopink!10}{\small\textit{Ours}} & \textbf{\Rorch{}} & \textbf{91.09}\sd{0.43} & \textbf{92.34}\sd{0.54} & \textbf{86.09}\sd{0.65} & \textbf{89.30}\sd{0.76} & \textbf{94.06}\sd{0.70} & \textbf{81.33}\sd{2.98} & \textbf{93.13}\sd{0.35} & \textbf{90.63}\sd{0.96} & \textbf{88.59}\sd{0.43} & \textbf{90.82}\sd{0.47} & \textbf{89.79}\sd{1.01} \\
\midrule\midrule
\multicolumn{2}{l}{\textit{(b) Out-of-distribution}} & \multicolumn{2}{c}{\textbf{TriviaQA}} & \multicolumn{2}{c}{\textbf{MuSiQue}} & \textbf{GPQA} & \textbf{MATH-Hard} & \textbf{SWE-Bench} & \textbf{WebShop} & \multicolumn{3}{c}{\textbf{OOD}} \\
\cmidrule(lr){3-4}\cmidrule(lr){5-6}\cmidrule(lr){7-7}\cmidrule(lr){8-8}\cmidrule(lr){9-9}\cmidrule(lr){10-10}\cmidrule(lr){11-13}
\textbf{Type} & \textbf{Method} & \footnotesize EM & \footnotesize F1 & \footnotesize EM & \footnotesize F1 & \footnotesize Acc. & \footnotesize Acc. & \footnotesize Resolved & \footnotesize SR & \footnotesize EM & \footnotesize F1 & \footnotesize Acc. \\
\midrule
\multirow{2}{*}{\small\textit{Backbone}} & Qwen3.5-9B & 44.53\sd{0.00} & 53.44\sd{0.06} & 39.22\sd{0.65} & 48.98\sd{0.92} & 61.41\sd{0.70} & 89.38\sd{0.43} & 15.78\sd{0.86} & 32.81\sd{0.78} & 41.88\sd{0.33} & 51.21\sd{0.46} & 49.84\sd{0.29} \\
 & DeepSeek-V4-Flash & 70.31\sd{1.66} & 80.94\sd{1.91} & 50.31\sd{0.43} & 58.59\sd{0.69} & 74.22\sd{1.66} & 92.50\sd{1.31} & 35.78\sd{0.86} & 65.31\sd{1.05} & 60.31\sd{0.76} & 69.77\sd{1.02} & 66.95\sd{0.58} \\
\cmidrule(lr){1-13}
\multirow{2}{*}{\small\textit{Training}} & SFT & 46.41\sd{0.70} & 58.91\sd{0.83} & 38.75\sd{0.70} & 50.78\sd{0.93} & 68.91\sd{0.65} & 87.97\sd{0.70} & 14.69\sd{0.65} & 32.97\sd{1.40} & 42.58\sd{0.28} & 54.85\sd{0.62} & 51.13\sd{0.51} \\
 & GRPO & 47.66\sd{0.55} & 58.67\sd{0.68} & 40.94\sd{1.52} & 50.08\sd{1.67} & 67.03\sd{0.86} & 87.81\sd{0.70} & 17.50\sd{0.43} & 35.94\sd{0.78} & 44.30\sd{0.90} & 54.38\sd{0.90} & 52.07\sd{0.30} \\
\cmidrule(lr){1-13}
\multirow{1}{*}{\small\textit{Search}} & AFlow & 90.31\sd{0.70} & 92.19\sd{1.04} & 77.19\sd{1.50} & 82.89\sd{1.75} & 75.78\sd{0.78} & 91.41\sd{0.00} & 28.28\sd{0.86} & 59.38\sd{0.78} & 83.75\sd{0.90} & 87.54\sd{1.02} & 63.71\sd{0.49} \\
\cmidrule(lr){1-13}
\multirow{2}{*}{\small\textit{Agent RL}} & AgentFlow & 88.91\sd{1.28} & 90.78\sd{1.65} & 79.06\sd{1.50} & 84.06\sd{1.72} & 79.53\sd{0.35} & 93.13\sd{0.65} & 39.38\sd{1.31} & 75.94\sd{0.65} & 83.98\sd{0.92} & 87.42\sd{1.19} & 71.99\sd{0.49} \\
 & FlowSteer & 92.50\sd{1.62} & 93.83\sd{1.89} & 79.69\sd{1.10} & 84.77\sd{1.43} & 83.91\sd{0.43} & 93.44\sd{0.43} & 37.97\sd{0.70} & 78.75\sd{1.40} & 86.09\sd{0.98} & 89.30\sd{1.19} & 73.52\sd{0.26} \\
\cmidrule(lr){1-13}
\multirow{2}{*}{\small\textit{Skill int.}} & SKILL0 & 91.25\sd{0.65} & 93.13\sd{0.74} & 79.22\sd{0.70} & 82.97\sd{0.88} & 81.25\sd{0.78} & 92.50\sd{1.05} & 36.88\sd{0.86} & 82.81\sd{1.56} & 85.23\sd{0.64} & 88.05\sd{0.57} & 73.36\sd{0.47} \\
 & PATS & 93.75\sd{0.00} & 94.53\sd{0.17} & 82.97\sd{0.65} & 86.64\sd{0.73} & 82.97\sd{0.65} & 93.44\sd{0.70} & 40.16\sd{0.70} & 87.97\sd{1.52} & 88.36\sd{0.33} & 90.59\sd{0.37} & 76.13\sd{0.42} \\
\cmidrule(lr){1-13}
\multirow{2}{*}{\small\textit{Self-play}} & SESA & 92.66\sd{0.43} & 94.06\sd{0.76} & 81.41\sd{0.86} & 85.55\sd{0.90} & 85.47\sd{0.70} & 92.81\sd{0.35} & 37.81\sd{1.05} & 77.66\sd{0.70} & 87.03\sd{0.51} & 89.81\sd{0.59} & 73.44\sd{0.53} \\
 & MetaSkill & 89.53\sd{1.05} & 92.27\sd{1.20} & 77.97\sd{0.86} & 81.64\sd{1.08} & 78.59\sd{0.70} & 91.56\sd{0.65} & 34.53\sd{0.86} & 72.19\sd{1.18} & 83.75\sd{0.21} & 86.96\sd{0.81} & 69.22\sd{0.55} \\
\midrule
\rowcolor{evopink!10}{\small\textit{Ours}} & \textbf{\Rorch{}} & \textbf{95.31}\sd{0.78} & \textbf{96.17}\sd{0.89} & \textbf{85.31}\sd{0.35} & \textbf{88.20}\sd{0.65} & \textbf{86.56}\sd{0.86} & \textbf{94.69}\sd{0.86} & \textbf{42.81}\sd{0.35} & \textbf{91.09}\sd{0.70} & \textbf{90.31}\sd{0.51} & \textbf{92.19}\sd{0.55} & \textbf{78.79}\sd{0.17} \\
\bottomrule
\end{tabular}}
\caption{Main results (five-run mean\,$\pm$\,std, \%), with benchmarks grouped by task domain. All methods share the frozen Workers, the training tasks, and the compute budget, each in its best configuration, and \Rorch{} is tested with zero skills. Each test set has 128 items (AIME 2026: 30); bold marks the best score in each benchmark column.}
\label{tab:main}
\end{table*}

\subsection{Skill Withdrawal}
\label{sec:sw}

Skill Withdrawal is the upward channel, through which the inner loop supplies the outer loop's evidence: it keeps each $C_t$ small by withdrawing a skill once the same rollout groups, read as a controlled evaluation, show that the Director performs as well without it (Figure~\ref{fig:method}, bottom right).

\noindent\textbf{Estimand and estimator.}
Because the hidden part is drawn at random, each rollout group is a randomized block \citep{rubin1974estimating,imbens2015causal}, and the difference of its part means is an unbiased (Neyman) estimator of the skill's effect:
\begin{equation}
\begin{gathered}
z_x=\tfrac1{n_v}\textstyle\sum_{i\in\mathcal V_x}y_i-\tfrac1{n_h}\sum_{i\in\mathcal H_x}y_i,\\
\mathbb E[z_x\mid x,k_x{=}k]=p_{K_x}-p_{K_x\setminus k}\eqqcolon\Delta_k(x),
\end{gathered}
\label{eq:contrast}
\end{equation}
where $k=k_x$.
Averaging over the tasks that test $k$ gives its benefit $\Delta_k=\mathbb E[\Delta_k(x)\mid k_x=k]$, which estimates its cost per retrieval in $C_t$ (Lemma~\ref{lem:decomp}).
The success rate mixes it with the curriculum:
\begin{equation}
\mathbb E[\bar y\mid x,k_x=k]=\underbrace{p_{K_x\setminus k}(x)}_{\smash{\text{curriculum}}}+\underbrace{\tfrac{n_v}{G}\,\Delta_k(x)}_{\smash{\text{skill}}}.
\label{eq:confound}
\end{equation}
Recursive Learnability pulls the curriculum term toward one half, so a threshold on $\bar y$ reads the curriculum rather than the skill, while $z_x$ removes it.

\noindent\textbf{Posterior and decisions.}
The contrasts accumulate into a Bayesian-bootstrap posterior per skill \citep{rubin1981bayesian}, and a Thompson draw tests first the skill closest to withdrawal \citep{william1933likelihood}:
\begin{equation}
\begin{gathered}
\Pi_k=\mathrm{Law}\big(\textstyle\sum_{x\in\mathcal D_k}w_xz_x\big),\ \ w\sim\mathrm{Dir}(\mathbf 1_{n_k}),\\
k_x=\arg\min_{k\in K_x}\ \tilde\Delta_k,\qquad \tilde\Delta_k\sim\Pi^{0}_k,
\end{gathered}
\label{eq:post}
\end{equation}
where $\mathcal D_k$ holds the $n_k$ records of skill $k$, and $\Pi^{0}_k$ adds one prior record so that untested skills are also drawn.
On this posterior, withdrawal and admission are Bayes decisions \citep{Berger_1985} under loss $\Delta^{+}$ for withdrawing, $\epsilon$ for keeping, and an asymmetric zero-one loss for admitting:
\begin{equation}
\begin{gathered}
\mathcal W_t=\big\{k:\ n_k\ge n_{\min},\ \mathbb E_{\Delta\sim\Pi_k}[\Delta^{+}]\le\epsilon\big\},\\
\mathcal N_t=\big\{k:\ n_k\ge n_{\min},\ q_k\le\alpha\big\},
\end{gathered}
\label{eq:decide}
\end{equation}
with $\Delta^{+}=\max(\Delta,0)$, $q_k=\Pi_k(\Delta\le0)$, $n_{\min}$ the minimum record, and $\alpha$ the admission risk, so each withdrawal costs at most $\epsilon$ in posterior expectation on the tasks on which its skill was tested.

\noindent\textbf{Failures to skills.}
The same posterior closes the outer loop: the least helpful skill is revised first, and new skills are distilled from groups where both parts fail (Appendix~\ref{app:loop}), so failures exposed by withdrawals become the next skills to internalize.

\noindent\textbf{Proposition 3.} \emph{Random hiding makes the paired contrast unbiased for a skill's benefit.}
\emph{Proof.} Appendix~\ref{app:proofs}, via randomization \citep{rubin1974estimating}.

\section{Experiments}

We evaluate \Rorch{} through the following research questions (RQs).
\textbf{RQ1:} How does \Rorch{} compare with the baselines in distribution and on held-out benchmarks?
\textbf{RQ2:} Does the trained Director transfer to other Worker backbones without retraining?
\textbf{RQ3:} Do both loops of Recursive Self-Play Orchestration contribute, and does the Director keep its accuracy at zero skills (Section~\ref{sec:paradigm})?
\textbf{RQ4:} Does Recursive Learnability choose the tasks that keep yielding a learning signal as skills are withdrawn (Section~\ref{sec:rl})?
\textbf{RQ5:} Does Skill Withdrawal remove each skill only once the Director no longer needs it (Section~\ref{sec:sw})?

\begin{table*}[p]
\centering
\setlength{\tabcolsep}{2.5pt}
\resizebox{\textwidth}{!}{
\begin{tabular}{l|cccccc|cccccc}
\toprule
\textbf{Variant} & \multicolumn{6}{c|}{\textbf{IID}} & \multicolumn{6}{c}{\textbf{OOD}} \\
\cmidrule(lr){2-7}\cmidrule(lr){8-13}
 & \textbf{HotpotQA} & \textbf{NQ-Open} & \textbf{MedQA} & \textbf{AIME} & \textbf{MBPP+} & \textbf{ALFWorld} & \textbf{TriviaQA} & \textbf{MuSiQue} & \textbf{GPQA} & \textbf{MATH} & \textbf{SWE} & \textbf{WebShop} \\
 & {\footnotesize Ans EM} & {\footnotesize Ans EM} & {\footnotesize Acc.} & {\footnotesize Acc.} & {\footnotesize Pass@1} & {\footnotesize SR} & {\footnotesize Ans EM} & {\footnotesize Ans EM} & {\footnotesize Acc.} & {\footnotesize Acc.} & {\footnotesize Resolved} & {\footnotesize SR} \\
\midrule
Qwen3.5-9B (direct prompting) & 57.81 & 23.59 & 70.94 & 48.67 & 78.59 & 47.19 & 44.53 & 39.22 & 61.41 & 89.38 & 15.78 & 32.81 \\
\midrule
\multicolumn{13}{l}{\small\textit{Recursive Self-Play Orchestration (Sec.~4.1)}} \\
$-$\,Recursive Self-Play Orchestration (GRPO only) & 86.09 & 82.97 & 88.44 & 59.33 & 86.88 & 73.44 & 91.41 & 78.91 & 80.78 & 92.03 & 35.00 & 71.72 \\
$-$\,Skills (no scaffold) & 86.88 & 83.59 & 89.22 & 64.00 & 88.13 & 77.81 & 92.34 & 80.63 & 82.66 & 92.34 & 37.66 & 76.09 \\
$-$\,Self-play (uniform picks) & 90.16 & 83.59 & 92.34 & 66.67 & 90.47 & 86.88 & 93.44 & 81.72 & 84.53 & 93.91 & 38.91 & 87.03 \\
\midrule
\multicolumn{13}{l}{\small\textit{Recursive Learnability (Sec.~4.2)}} \\
$-$\,Recursive Learnability (failure reward) & 87.97 & 82.66 & 90.47 & 64.00 & 90.00 & 84.06 & 92.66 & 81.09 & 83.44 & 92.81 & 38.13 & 83.13 \\
Learnability on the current scaffold & 90.47 & 84.38 & 92.03 & 70.67 & 91.25 & 86.88 & 94.69 & 83.13 & 85.47 & 93.59 & 40.00 & 85.94 \\
Dynamic sampling (DAPO) & 88.13 & 83.59 & 92.50 & 68.67 & 91.09 & 87.50 & 94.69 & 82.97 & 84.84 & 93.91 & 39.53 & 87.81 \\
\midrule
\multicolumn{13}{l}{\small\textit{Skill Withdrawal (Sec.~4.3)}} \\
$-$\,Skill Withdrawal (scaffold fixed) & 88.13 & 82.97 & 90.47 & 66.67 & 88.75 & 77.81 & 92.34 & 81.25 & 83.28 & 92.81 & 38.13 & 76.09 \\
Success-rate threshold (PATS) & 90.00 & 84.53 & 91.88 & 74.67 & 90.31 & 87.81 & 93.59 & 82.97 & 84.69 & 94.06 & 39.53 & 87.66 \\
Preset budget (SKILL0) & 89.69 & 84.22 & 91.25 & 70.67 & 90.63 & 86.09 & 93.75 & 82.66 & 84.38 & 93.75 & 39.38 & 85.00 \\
Validation gap (SLIM) & 90.63 & 85.00 & 93.28 & 72.67 & 91.56 & 88.28 & 94.84 & 83.28 & 85.31 & 93.59 & 40.47 & 89.22 \\
\midrule
\rowcolor{evopink!10}\textbf{\Rorch{} (Full)} & \textbf{91.09} & \textbf{86.09} & \textbf{94.06} & \textbf{81.33} & \textbf{93.13} & \textbf{90.63} & \textbf{95.31} & \textbf{85.31} & \textbf{86.56} & \textbf{94.69} & \textbf{42.81} & \textbf{91.09} \\
\bottomrule
\end{tabular}}
\caption{Ablation of \Rorch{}, tested with zero skills (five-run mean, \%). The first row of each group removes the whole component and the other rows change one part (Appendix~\ref{app:exp}); free-form QA benchmarks report Ans EM, and the Qwen3.5-9B and \Rorch{} rows repeat their scores in Table~\ref{tab:main}.}
\label{tab:ablate}
\end{table*}

\begin{figure*}[p]
\centering
\includegraphics[width=\textwidth]{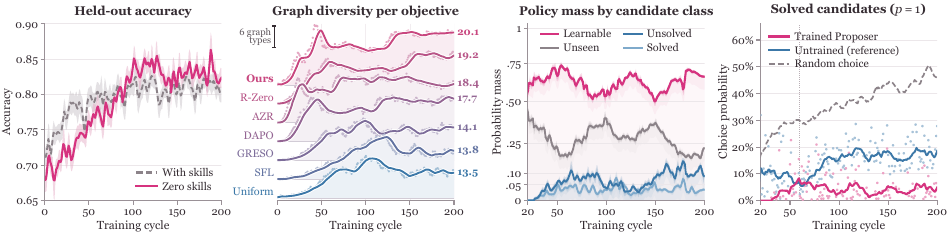}\\[1pt]
\makebox[\textwidth][l]{\small\hspace*{15.88pt}\makebox[96pt]{(a) Held-out accuracy}\hspace*{15.66pt}\makebox[96pt]{(b) Graph diversity}\hspace*{24.26pt}\makebox[96pt]{(c) Policy mass}\hspace*{15.66pt}\makebox[96pt]{(d) Solved candidates}}
\par\vspace{6pt}
\includegraphics[width=\textwidth]{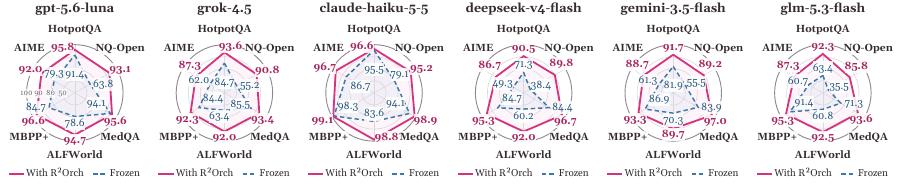}\\[2pt]
\includegraphics[width=\textwidth]{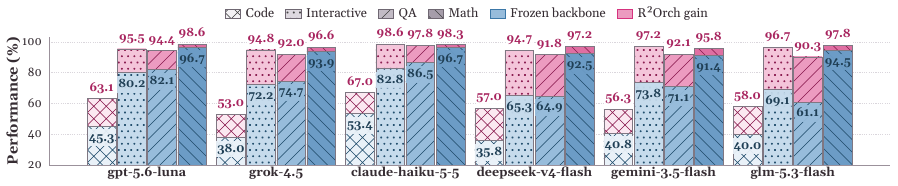}\\[1pt]
{\small (e) Backbone transfer: IID (radars) and OOD (bars)}
\caption{Training dynamics and backbone transfer. (a) Held-out accuracy with skills and at zero skills. (b) Graph types over the six IID datasets per Proposer objective (final counts at the curve ends). (c) The Proposer's probability mass on unseen, unsolved ($p=0$), learnable ($0<p<1$), and solved ($p=1$) candidates by previous success rate $p$ (power-scaled axis). (d) Probability of choosing a solved candidate as a moving average over 10 cycles (dots: single cycles) for the trained Proposer, the untrained model with the same prompt, and random choice. (e) Backbone transfer: the Director trained with Qwen3.5-9B Workers is used unchanged at zero skills with six other backbones. Radars: IID scores with \Rorch{} (solid) and without it (dashed), the radius linear over 20--80 (inner half) and 80--100 (outer half). Bars: OOD scores per task domain, the frozen backbone in blue and \Rorch{}'s gain in pink (Code: SWE-Bench; Interactive: WebShop; QA: mean of TriviaQA, MuSiQue, and GPQA; Math: MATH-Hard).}
\label{fig:transfer}
\end{figure*}

\begin{figure*}[t]
\centering
\sbox{\objtabbox}{\setlength{\tabcolsep}{1.5pt}\renewcommand{\arraystretch}{1.15}\resizebox{0.709\textwidth}{!}{
\begin{tabular}[b]{lcccccc}
\toprule
\textbf{Proposer objective} & \textbf{HotpotQA} & \textbf{NQ-Open} & \textbf{MedQA} & \textbf{AIME} & \textbf{MBPP+} & \textbf{ALFWorld} \\
 & {\footnotesize Ans EM} & {\footnotesize Ans EM} & {\footnotesize Acc.} & {\footnotesize Acc.} & {\footnotesize Pass@1} & {\footnotesize SR} \\
\midrule
Uniform sampling & 90.16 & 83.59 & 92.34 & 66.67 & 90.47 & 86.88 \\
Dynamic sampling (DAPO) & 88.13 & 83.59 & 92.50 & 68.67 & 91.09 & 87.50 \\
Selective rollout (GRESO) & 88.44 & 83.28 & 91.56 & 68.00 & 90.31 & 85.78 \\
Learnability, current scaffold (SFL) & 90.47 & 84.38 & 92.03 & 70.67 & 91.25 & 86.88 \\
Solve-rate reward (Absolute Zero) & 90.78 & 84.53 & 92.34 & 77.33 & 92.19 & 88.13 \\
Uncertainty reward (R-Zero) & 90.31 & 84.38 & 92.50 & 70.00 & 91.72 & 87.34 \\
\midrule
\rowcolor{evopink!10}\textbf{Recursive Learnability (ours)} & \textbf{91.09} & \textbf{86.09} & \textbf{94.06} & \textbf{81.33} & \textbf{93.13} & \textbf{90.63} \\
\bottomrule
\end{tabular}}}
\begin{minipage}[b]{0.709\textwidth}\centering
\usebox{\objtabbox}\\[3pt]
{\small (a) IID scores per Proposer objective}
\end{minipage}\hfill
\begin{minipage}[b]{\dimexpr\textwidth-0.709\textwidth-4pt\relax}\centering
\includegraphics[height=\dimexpr\ht\objtabbox+\dp\objtabbox-4.5pt\relax]{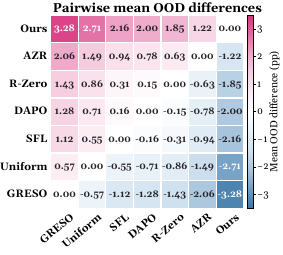}\\[3pt]
{\small (b) Pairwise OOD differences}
\end{minipage}
\par\vspace{7pt}
\includegraphics[width=\textwidth]{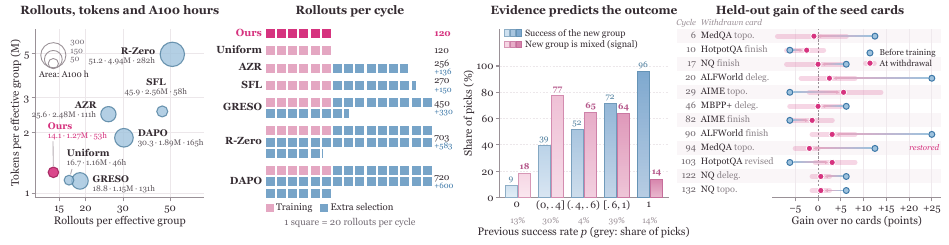}\\[1pt]
\makebox[\textwidth][l]{\small\hspace*{12.41pt}\makebox[90pt]{(c) Per effective group}\hspace*{20.51pt}\makebox[90pt]{(d) Per cycle}\hspace*{22.31pt}\makebox[90pt]{(e) Evidence and outcome}\hspace*{41.01pt}\makebox[90pt]{(f) Skill withdrawals}}

\caption{Proposer objectives, selection cost, and skill withdrawals (RQ4, RQ5); only the Proposer's objective changes in (a)--(d) (AZR: Absolute Zero). (a) IID scores. (b) Row minus column of the mean OOD score. (c) Rollouts and tokens per effective group; bubble area: A100 hours. (d) Training (pink) and selection (blue) rollouts per cycle. (e) Seen-task picks over cycles 50--199 by previous success rate $p$: realized success and share of mixed new groups. (f) The seed cards' held-out gain over no cards, before training and around each withdrawal (95\% CI).}
\label{fig:objective}
\end{figure*}

\noindent\textbf{Setup.}
Six in-distribution (IID) benchmarks supply the training tasks and IID tests, and six out-of-distribution (OOD) benchmarks are held out.
The baselines span direct prompting, backbone training, workflow search, RL-trained orchestration, skill internalization, and self-play with skill evolution, under the shared protocol of Table~\ref{tab:main} (Appendix~\ref{app:exp}).

\noindent\textbf{Main results (RQ1).}
\Rorch{} is best on all twelve benchmarks against six families of baselines, and its lead holds on the held-out half at zero skills (Table~\ref{tab:main}).
Training the backbone alone changes little, while orchestration lifts the frozen Workers far higher, so the gains come from how the Workers are organized.
The other orchestrators learn from signals that stop tracking the Director: fixed task pools (workflow search, RL-trained orchestration), success rates or budgets (skill internalization), and frontiers measured with skills in context (self-play).
\Rorch{} replaces these signals with Recursive Learnability and Skill Withdrawal, so its lead is largest on AIME 2026, WebShop, and ALFWorld, where the ablations cost most (Table~\ref{tab:ablate}).

\noindent\textbf{Backbone transfer (RQ2).}
The Director trained with Qwen3.5-9B Workers improves all six other backbones on every benchmark without retraining (Figure~\ref{fig:transfer}e), since the roles and hand-offs it builds fit any Worker.
Weaker backbones gain the most, narrowing the IID range across backbones from 25.7 to 6.4 points, because the graph supplies the planning and checking they lack.
On OOD benchmarks, the mean gain is largest on interactive tasks, whose many steps compound each hand-off, and smallest on math, which is already near its ceiling.

\noindent\textbf{Recursive Self-Play Orchestration (RQ3).}
Both loops are needed (Table~\ref{tab:ablate}): skills give the Director the early successes that make hard tasks learnable, and the Proposer keeps groups away from uniform outcomes.
Since most of each group hides the candidate skill (Eq.~\eqref{eq:outer}), updates teach the Director to succeed without it, and zero-skill accuracy overtakes the accuracy with skills midway (Figure~\ref{fig:transfer}a).
A skill has thus stopped helping before it leaves, keeping the outer costs of Eq.~\eqref{eq:decomp} small as the inner gains of self-play accumulate (Proposition~1).

\noindent\textbf{Recursive Learnability (RQ4).}
Recursive Learnability picks the tasks that Proposition~2 expects to gain most, those whose rollouts still disagree with the candidate skill withheld.
New groups on tasks of intermediate success rate reach mixed outcomes, and thus a nonzero GRPO signal, about four times as often as at $p=0$ or $p=1$ (Figure~\ref{fig:objective}e), and the trained Proposer learns to skip solved candidates (Figure~\ref{fig:transfer}c,d).
Learnability measured with the skill in context picks tasks that collapse once the skill leaves, so it trails on every IID benchmark and the mean OOD score, as does every other objective (Table~\ref{tab:ablate}, Figure~\ref{fig:objective}a,b).
Since the hidden part already measures this score, selection costs no extra rollouts and the fewest per effective group (Figure~\ref{fig:objective}c,d), and these groups lead the Director to build the most distinct graph types (Figure~\ref{fig:transfer}b).

\noindent\textbf{Skill Withdrawal (RQ5).}
Skill Withdrawal removes a skill once it no longer helps the current Director, as it reads the skill's own effect, not the success rate.
The success rate mixes that effect with the curriculum that Recursive Learnability shapes (Eq.~\eqref{eq:confound}), so a success-rate threshold misreads the skill and trails the paired contrast, as do a preset budget and a validation gap (Proposition~3, Table~\ref{tab:ablate}).
All 12 withdrawals pass Eq.~\eqref{eq:decide} on the card's own contrasts, with posterior $\mathbb E[\Delta^{+}]$ at most 0.019, and around each the seed cards no longer raise held-out accuracy (Figure~\ref{fig:objective}f).
A fixed scaffold costs most at zero skills, as the Director keeps leaning on skills that never leave its context throughout training.

\section{Conclusion}

\Rorch{} nests two loops: Recursive Learnability picks tasks learnable without the candidate skill, and Skill Withdrawal retires skills that stop helping.
It leads on all twelve benchmarks at zero skills.

\section*{Limitations}
\Rorch{} starts from initial skills that we customize by hand for each dataset, and the withdrawals, revisions, and distillations that follow build on them, so deriving the initial skills automatically for a new domain remains an open problem.

\section*{Ethical considerations}
\Rorch{} is trained and evaluated on public benchmarks that we use for research evaluation as their creators intend, and the study involves no human participants, annotators, or personal data. ALFWorld and WebShop are simulated environments, and the code benchmarks run in their standard evaluation harnesses. Agents that orchestrate tools can be misused like any agent system, and a self-improving orchestrator shares this risk, so we release the code for inspection and recommend running it in a sandboxed environment.
\section*{Acknowledgements}
We used AI assistants to polish the language and to help with \LaTeX{} typesetting; all other work, including the experiments and the first drafts, is the authors' own, and the authors take full responsibility for the content of the paper.

\bibliography{refs}
\clearpage
\appendix
\makeatletter\setlength{\@fptop}{0pt}\setlength{\@fpsep}{12pt plus 2pt}\makeatother

\section{Theory and Proofs}
\label{app:proofs}

This appendix analyzes the two channels of \Rorch{}: outcome diversity for the curriculum and randomized evidence for skill withdrawal. We derive the policy-space progress calculation, the rollout estimators, and the posterior decisions, then connect them through round accounting. Proposition~1 rests on Lemmas~\ref{lem:mixed}--\ref{lem:twotime}, Proposition~2 on Lemmas~\ref{lem:tilt}--\ref{lem:energy} and~\ref{lem:proposer}, and Proposition~3 on Lemmas~\ref{lem:random} and~\ref{lem:bayes}; the final subsection collects the three proofs, and Table~\ref{tab:notation} the notation.

\begin{table}[!htb]
\centering\small
\setlength{\tabcolsep}{4pt}
\begin{tabular}{@{}ll@{}}
\toprule
Symbol & Meaning\\
\midrule
$\pi_\theta$, $\rho_\phi$, $\rho_0$ & Director, Proposer, Proposer reference\\
$\rho$, $\mathcal C_t$ & evaluation distribution, candidate list\\
$\mathcal S_t$, $K_x$, $k_x$ & scaffold, skills retrieved for $x$, skill under test\\
$G=n_h+n_v$ & rollouts per group (hidden, visible)\\
$\mathcal H_x$, $\mathcal V_x$ & parts reading $K_x\setminus k_x$ and $K_x$\\
$y_i$, $u_i$ & correctness; correctness less Worker cost\\
$p^{+}$, $p^{-}$ & success probability under $K_x$, $K_x\setminus k_x$\\
$\ell(p)$, $\ell_{\mathrm{rec}}$ & learnability $4p(1-p)$; its value at $p^-$\\
$R_P$, $\tilde p_x$ & Proposer reward; shrunk success rate\\
$z_x$, $\Delta_k(x)$ & paired contrast; skill effect on task $x$\\
$\Pi_k$, $\Pi^{0}_k$ & posterior of $\Delta_k$; with one prior record\\
$\mathcal W_t$, $\mathcal N_t$ & skills withdrawn, admitted in round $t$\\
$J_{\mathcal S}$, $\Gamma_t$, $C_t$ & success rate under $\mathcal S$; inner gain; outer cost\\
$\beta$, $\tau$ & Director step size; Proposer KL weight\\
\bottomrule
\end{tabular}
\caption{Notation of the main text.}
\label{tab:notation}
\end{table}

\subsection{Conditional setup and policy likelihoods}
\label{app:theory-setup}

Fix a task $x$, a scaffold, a retrieved context $K$, and a candidate skill $k\in K$. Write $K^-=K\setminus\{k\}$, $p^+=p_K(x;\theta)$, and $p^-=p_{K^-}(x;\theta)$. The success probability averages both Director sampling and Worker execution. The two parts have fixed positive sizes $n_h,n_v$, with $G=n_h+n_v$. During one group, the policy, candidate, contexts, and pre-group statistics are held fixed. Conditional on the assigned contexts, different rollouts use independent randomness. Task rewards average candidate-skill selection whenever several skills can be tested for the same task. Gradients hold the task and context sampling laws fixed and assume the usual measurability and integrability conditions.

For an execution that returns intermediate Worker feedback, let $\omega$ denote its complete record and $h_t$ the history before Director action $a_t$. Frozen Worker kernels are independent of $\theta$, so the likelihood score is
\begin{equation}
 S_\theta(\omega)
 =\nabla_\theta\log P_\theta(\omega)
 =\sum_t\nabla_\theta\log\pi_\theta(a_t\mid h_t).
 \label{eq:app-score}
\end{equation}
Thus $\mathbb E[S_\theta]=0$ and $\nabla_\theta\mathbb E[u]=\mathbb E[uS_\theta]$ for a reward with no direct parameter dependence. Multiple editing histories may produce the same terminal graph; their individual action likelihoods are not the marginal likelihood of that graph. The group-gradient results use the record score in Eq.~\eqref{eq:app-score}. A terminal-graph score gives the same value gradient when the conditional reward mean given that graph is independent of $\theta$. Otherwise the complete record is required, for example when editing histories with different costs reach the same graph.

The policy-space tilt below considers a choice $g$ followed by a fixed execution kernel, so its stochastic extension leaves Worker randomness outside the Director's control. We use a fixed evaluation distribution $\rho$ for round accounting. The training curriculum $\rho_\phi$ and the distribution of tasks recorded for a particular skill need not equal $\rho$.

In Eq.~\eqref{eq:nested}, the Director KL is understood as averaged over the fixed task and context distribution. Its moving reference $\pi_{\theta_t}$ defines an ideal proximal response; the base-model anchor in the training loss is a separate regularizer. The Proposer instead uses the fixed reference $\rho_0$ in both its KL term and the trust region on its pair tilt (Appendix~\ref{app:impl}), so its response is proximal to $\rho_0$. 

\subsection{KL tilting and controllable progress}

\begin{lemma}[KL-tilt progress]\label{lem:tilt}
Let $\pi$ be a reference distribution and $F(g)$ a bounded, parameter-independent score. For $\beta>0$, the unique maximizer, on the support of $\pi$, of
$\mathbb E_{\pi'}[F]-\beta^{-1}\mathrm{KL}(\pi'\|\pi)$ is
\begin{equation}
 \pi_\beta(g)=\frac{\pi(g)e^{\beta F(g)}}{Z(\beta)},
 \qquad Z(\beta)=\mathbb E_\pi[e^{\beta F}].
 \label{eq:app-general-tilt}
\end{equation}
Writing $\psi=\log Z$, $\mu=\mathbb E_\pi F$, and $V=\mathrm{Var}_\pi(F)$ gives
\begin{equation}
 \begin{aligned}
 \psi(\beta)&=\beta\mu+\tfrac12\beta^2V+O(\beta^3),\\
 \mathbb E_{\pi_\beta}F-\mu&=\beta V+O(\beta^2),\\
 \mathrm{KL}(\pi_\beta\|\pi)&=\tfrac12\beta^2V+O(\beta^3).
 \end{aligned}
 \label{eq:app-local-progress}
\end{equation}
For fixed $V>0$, a small KL budget $\delta$ yields optimal score gain $\sqrt{2\delta V}+O(\delta)$.
\end{lemma}
\begin{proof}
Substitution of the tilted density gives, for every admissible $\pi'$,
\begin{equation}
 \mathbb E_{\pi'}F-\beta^{-1}\mathrm{KL}(\pi'\|\pi)
 =\beta^{-1}\{\log Z-\mathrm{KL}(\pi'\|\pi_\beta)\}.
 \label{eq:app-variational}
\end{equation}
Nonnegativity of KL proves optimality and uniqueness. Boundedness permits differentiation of $Z$: $\psi'(\beta)=\mathbb E_{\pi_\beta}F$ and $\psi''(\beta)=\mathrm{Var}_{\pi_\beta}(F)$. Expanding at zero proves the first two identities. Since the log density ratio is $\beta F-\psi(\beta)$, its expectation is $\beta\psi'(\beta)-\psi(\beta)$, proving the KL expansion. For $V>0$, inversion gives $\beta=\sqrt{2\delta/V}+O(\delta)$. The variational inequality also shows that this tilt maximizes expected score among distributions with its KL cost. Substitution proves the budget statement. If $V=0$, the score is constant on the available support and no such tilt improves it.
\end{proof}

For a binary outcome determined by $g$, take $F=r$ and $p=\mathbb E_\pi r$. Then
\begin{equation}
 \begin{aligned}
 Z(\beta)&=1-p+pe^\beta,\qquad p_\beta=\frac{pe^\beta}{Z(\beta)},\\
 p_\beta-p&=\frac{p(1-p)(e^\beta-1)}{1+p(e^\beta-1)}.
 \end{aligned}
 \label{eq:app-bernoulli}
\end{equation}
Consequently $\psi''(0)=p(1-p)$ is also the Fisher information of this one-dimensional exponential family. At a common small tilt, gains rank by $p(1-p)$ to first order; at a common small KL budget, they scale with its square root. At a finite tilt the ranking can change: at $\beta=\log 2$, $p=0.3$ yields gain $0.21/1.3$, whereas $p=0.6$ yields the smaller gain $0.24/1.6$ despite its larger variance. Gain per unit KL cost is $2/\beta+O(1)$.

\paragraph{Stochastic Workers.}
Let $Q(g)=\mathbb E[Y\mid g]$ for a fixed Worker execution kernel and $Y\in\{0,1\}$. Optimizing expected correctness over graph distributions applies Lemma~\ref{lem:tilt} with $F=Q$, not with a sampled $Y$. The controllable graph-level variance is
\begin{equation}
 \begin{aligned}
 V_{\mathrm{ctrl}}&=\mathrm{Var}_\pi(Q),\\
 p(1-p)&=V_{\mathrm{ctrl}}+
       \mathbb E_\pi[Q(1-Q)].
 \end{aligned}
 \label{eq:app-worker-noise}
\end{equation}
The second identity follows by conditioning $Y^2=Y$ on $g$, or by total variance. A Director-only update reweights graphs and keeps the Worker kernel, whereas a joint tilt of graph and execution outcomes by $e^{\beta Y}$ would change that kernel. In particular, if $Q(g)=1/2$ for every graph, correctness variance is $1/4$ but no graph redistribution improves success. Recursive Learnability measures outcome diversity under stochastic execution, which equals its controllable part for graph-determined outcomes, the setting of Proposition~2.

A local statement also covers feedback-dependent records and restricted policy classes. Let $b=\mathbb E[(Y-p)S_\theta]$ and $I=\mathbb E[S_\theta S_\theta^\top]$. Under a differentiable local KL expansion, maximizing $b^\top d$ subject to $d^\top I d/2\le\delta$ gives
\begin{equation}
 \sqrt{2\delta\,b^\top I^\dagger b},
 \qquad b^\top I^\dagger b\le p(1-p).
 \label{eq:app-tangent}
\end{equation}
Indeed, a null direction of $I$ has zero score almost surely and is orthogonal to $b$. Cauchy--Schwarz in the range of $I$ gives the optimum. Moreover $S_\theta^\top I^\dagger b$ is the least-squares projection of $Y-p$ onto the score span, so its variance cannot exceed $\mathrm{Var}(Y)$. This identifies the accessible part of outcome diversity for any differentiable policy class.

For the cost-adjusted reward $u=Y-\lambda c$, define $U(g)=Q(g)-\lambda\mathbb E[c\mid g]$. Tilting by $U$ changes correctness at the initial rate $\mathrm{Cov}_\pi(Q,U)=\mathrm{Var}_\pi(Q)-\lambda\,\mathrm{Cov}_\pi(Q,\bar c)$ with $\bar c(g)=\mathbb E[c\mid g]$, which differs from $\mathrm{Var}_\pi(Q)$ by at most $\lambda\,(\mathrm{Var}_\pi(Q)\,\mathrm{Var}_\pi(\bar c))^{1/2}$, small for the weight $\lambda$ of Table~\ref{tab:hyper}.

\begin{lemma}[Equal tilt]\label{lem:equal}
For finitely many tasks with fixed weights $a_j>0$ and variances $V_j>0$, the quadratic-budget problem
\begin{equation}
 \max_{\beta_j\ge0}\sum_j a_j\beta_jV_j
 \quad\text{subject to}\quad
 \tfrac12\sum_j a_j\beta_j^2V_j\le\delta
 \label{eq:app-shared-tilt}
\end{equation}
has the common solution $\beta_j=\sqrt{2\delta/\sum_j a_jV_j}$.
\end{lemma}
\begin{proof}
Cauchy--Schwarz bounds the squared objective by
$(\sum_j a_j\beta_j^2V_j)(\sum_j a_jV_j)$, with equality exactly when all positive-variance tasks have the same tilt. Saturating the budget gives the stated value. Zero-variance tasks add no gain or cost.
\end{proof}
This allocation result for independently adjustable policy distributions motivates one advantage scale across tasks: the batch whitening of Appendix~\ref{app:impl} rescales every task by the same factor, whereas dividing by each task's standard deviation would give larger multipliers to low-variance tasks.

\subsection{Advantage energy, collapse, and shrinkage}

\begin{lemma}[Advantage energy]\label{lem:energy}
Let $n\ge2$ independent outcomes have distribution $\mathrm{Bernoulli}(p)$ and let $A_i=Y_i-\bar Y$. Then
\begin{equation}
 \begin{aligned}
 \mathbb E\sum_i A_i^2&=(n-1)p(1-p),\\
 \Pr\!\left(\sum_i A_i^2=0\right)&=p^n+(1-p)^n.
 \end{aligned}
 \label{eq:app-energy}
\end{equation}
Thus $\ell(p)=4p(1-p)$ is $4/(n-1)$ times the expected advantage energy, and a realized group collapses exactly when its outcomes agree.
\end{lemma}
\begin{proof}
For $B=\sum_iY_i$, binary outcomes give $\sum_iA_i^2=B-B^2/n$. Since $\mathbb EB=np$ and $\mathbb EB^2=np(1-p)+n^2p^2$, taking expectations proves the first identity. The sum vanishes exactly at $B\in\{0,n\}$, giving the second identity.
\end{proof}
For the actual hidden part, $n=n_h$, and the hypothetical homogeneous group in Eq.~\eqref{eq:recl} has $n=G$. Population learnability vanishes only at $p\in\{0,1\}$, while a finite group can collapse at any $p$: even at $p=1/2$, with probability $2^{1-n}$.
For $n>1$, differentiating the collapse probability gives
$n[p^{n-1}-(1-p)^{n-1}]$, which is negative below one half and positive above it. Thus the same population frontier that maximizes learnability minimizes homogeneous-group collapse probability. The hidden part and the full group consequently have different finite-sample collapse probabilities even when their success probabilities coincide.

The mixed group has a different energy. Let $V_-=p^-(1-p^-)$ and $V_+=p^+(1-p^+)$. Conditional independence gives
\begin{equation}
 \begin{aligned}
 \mathbb E\sum_i(Y_i-\bar Y)^2
 &=\left(1-\tfrac1G\right)(n_hV_-+n_vV_+)\\
 &\quad+\tfrac{n_hn_v}{G}(p^+-p^-)^2.
 \end{aligned}
 \label{eq:app-mixed-energy}
\end{equation}
To verify this, expand $\sum_iY_i^2-G\bar Y^2$ and separate within-context variances from the squared differences between context means. Its collapse probability is
$(p^-)^{n_h}(p^+)^{n_v}+(1-p^-)^{n_h}(1-p^+)^{n_v}$.
The extra energy term measures dependence on the skill rather than post-withdrawal learning potential, which is why Eq.~\eqref{eq:rp} reads the hidden part. If advantages use $u$, the same expansion concerns reward variance, and correctness agreement need not make them zero when costs differ.

\paragraph{Meaning of the Beta reward.}
Condition on a pre-group running mean $\bar p\in(0,1)$ and strength $m>0$. A proper working prior is $p\sim\mathrm{Beta}(a,b)$ with $a=m\bar p$, $b=m(1-\bar p)$. For a fixed context with $s$ successes in $n$ independent trials, set $a'=a+s$, $b'=b+n-s$, and $N=n+m$. Multiplication of prior and Bernoulli likelihood gives the posterior $\mathrm{Beta}(a',b')$, with
\begin{equation}
 \begin{aligned}
 \tilde p&=a'/N,\qquad \mathrm{Var}(p\mid y)=\frac{\tilde p(1-\tilde p)}{N+1},\\
 \mathbb E[\ell(p)\mid y]&=\frac{N}{N+1}\ell(\tilde p).
 \end{aligned}
 \label{eq:app-beta}
\end{equation}
The last identity follows from $\mathbb E[p(1-p)]=a'b'/[N(N+1)]$. Hence Eq.~\eqref{eq:rp} uses the posterior-mean \emph{plug-in}, which exceeds the posterior expectation of learnability by $4\mathrm{Var}(p\mid y)$. A running mean estimated from data gives an empirical-Bayes interpretation conditional on that mean. 

The plug-in also has a precise frequentist interpretation when the running mean is fixed before sampling. Let
$\mu_n=(np+m\bar p)/(n+m)$. Since $\mathbb E\tilde p=\mu_n$ and $\mathrm{Var}(\tilde p)=np(1-p)/(n+m)^2$, expanding the quadratic learnability function gives
\begin{equation}
 \mathbb E[\ell(\tilde p)]
 =\ell(\mu_n)-\frac{4np(1-p)}{(n+m)^2}.
 \label{eq:app-plugin-bias}
\end{equation}
There are therefore two separate effects: shrinkage changes the center, while finite sampling subtracts a variance term. The identity holds before the outcome-dependent branch selection below.

For paired groups, the stated branching can be written as follows: $c_x$ indicates agreement of all $G$ outcomes; on a noncollapsed group, $h_x=1$ when the hidden part has no success and the visible part does. Use $\mathcal A_x=\mathcal V_x$ in this transfer branch and $\mathcal A_x=\mathcal H_x$ otherwise. A no-skill group uses all its rollouts. The visible branch estimates visible-context success $p^+$ and supplies a discounted transfer incentive. The collapse penalty and the compute factor further shape the operational reward, whose expectation $\bar R_P$ enters Eq.~\eqref{eq:gibbs}.

There is nevertheless a direct large-group connection in the ordinary hidden-success branch. Fix $0<p^-<1$, let both part sizes increase, and take fixed $m$ or $m/n_h\to0$. Consistency of the hidden sample mean gives $\tilde p_x\to p^-$ in probability. The hidden-zero event has probability $(1-p^-)^{n_h}\to0$. Full-group collapse is contained in the event that the hidden outcomes all agree, whose probability also tends to zero. Therefore, with the compute factor omitted, the bounded reward in Eq.~\eqref{eq:rp} converges to $\ell(p^-)$ in probability and in expectation.

The exceptional branch still matters. At $p^-=0$ and $0<p^+<1$, the hidden part keeps failing while the visible part includes a success with high probability. With a consistent visible estimate, the reward approaches $\gamma\ell(p^+)$, so the transfer term keeps a discounted incentive for assisted practice on tasks whose hidden-context learnability is zero.

\subsection{Mixed-context gradients and the last skill}

\begin{lemma}[Mixed-group gradient]\label{lem:mixed}
Condition on an assignment of exactly $n_h$ hidden and $n_v$ visible contexts. Let $S_i$ be each rollout's likelihood score and $u_i$ its integrable reward. With independent rollouts conditional on this assignment and a common group mean,
\begin{equation}
 \mathbb E\left[\frac{\sum_i(u_i-\bar u)S_i}{G-1}\right]
 =\frac{n_h}{G}\nabla v_{K^-}
  +\frac{n_v}{G}\nabla v_K.
 \label{eq:app-mixed-gradient}
\end{equation}
If each part is centered separately, the expectation of the unnormalized sum becomes
\begin{equation}
 (n_h-1)\nabla v_{K^-}+(n_v-1)\nabla v_K.
 \label{eq:app-part-gradient}
\end{equation}
\end{lemma}
\begin{proof}
Write $\mathbf K$ for the complete context assignment. For $j\ne i$, conditional independence and the zero-mean score identity imply
$\mathbb E[u_jS_i\mid\mathbf K]=\mathbb E[u_j\mid\mathbf K]\mathbb E[S_i\mid\mathbf K]=0$.
For $j=i$, $\mathbb E[u_iS_i\mid\mathbf K]=\nabla v_{K_i}$. Therefore
\begin{equation}
 \mathbb E[(u_i-\bar u)S_i\mid\mathbf K]
 =\left(1-\tfrac1G\right)\nabla v_{K_i}.
 \label{eq:app-baseline-step}
\end{equation}
Summing and dividing by $G-1$ proves Eq.~\eqref{eq:app-mixed-gradient}. It also follows from the exact leave-one-out identity
$u_i-\bar u=(1-1/G)(u_i-\bar u_{-i})$.
Repeating the calculation within a part of size $n_c$ gives a total $(n_c-1)\nabla v_c$, proving Eq.~\eqref{eq:app-part-gradient}. Averaging over assignments preserves both conclusions.
\end{proof}

If the last-skill case $K=\{k\}$ uses separate centering, normalizing Eq.~\eqref{eq:app-part-gradient} by $G-2$ gives the mixture
\begin{equation}
 q'_x=\frac{n_h-1}{G-2}\delta_{\varnothing}
       +\frac{n_v-1}{G-2}\delta_{\{k\}}.
 \label{eq:app-last-skill}
\end{equation}
For $G>2$ and $n_h>n_v\ge1$, its hidden weight exceeds $n_h/G$ by $(n_h-n_v)/[G(G-2)]$. Keeping the divisor $G-1$ instead gives $(G-2)/(G-1)$ times this mixture gradient: the $3/2$ split yields weights $2:1$ and relative scale $3/4$. A singleton part contributes no centered gradient. Once no skill is retrieved, all $G$ rollouts share the empty context and ordinary whole-group centering applies.

These identities describe the on-policy reward gradient; PPO clipping, base-model anchoring, batch whitening, token averaging, and advantage-based filtering turn it into the training surrogate of Appendix~\ref{app:impl} on the same context mixture.

\subsection{Round accounting and interacting withdrawals}

\begin{lemma}[Round decomposition]\label{lem:decomp}
Fix $\rho$ and define
$J_{\mathcal S}(\theta)=\mathbb E_{x\sim\rho}[p_{K_x(\mathcal S)}(x;\theta)]$.
For any finite sequence of policies and scaffolds, define $\Gamma_t$ and $C_t$ as in Eq.~\eqref{eq:decomp}. Then
\begin{equation}
 J_{\mathcal S_T}(\theta_T)
 =J_{\mathcal S_0}(\theta_0)+\sum_{t<T}(\Gamma_t-C_t).
 \label{eq:app-round-identity}
\end{equation}
If an outer change removes distinct skills $k_1,\ldots,k_m$ without changing or refilling the remaining retrieved context, put
$K_x^j=K_x\setminus\{k_1,\ldots,k_j\}$. At $\theta_{t+1}$ its cost is
\begin{equation}
 C_t=\sum_{j=1}^m\mathbb E_\rho\left[
       p_{K_x^{j-1}}(x)-p_{K_x^j}(x)\right].
 \label{eq:app-sequential-cost}
\end{equation}
A term is zero on tasks not retrieving $k_j$.
\end{lemma}
\begin{proof}
Add and subtract $J_{\mathcal S_t}(\theta_{t+1})$ in each round's total change to obtain $\Gamma_t-C_t$, and telescope over rounds. For pure removals, telescope again over $K_x^0,\ldots,K_x^m$ before taking the task expectation. Both calculations are algebraic, so they hold for any sequence of policies and scaffolds.
\end{proof}

The sequential effects in Eq.~\eqref{eq:app-sequential-cost} can differ from single-removal effects. For two redundant skills $a,b$ with success probabilities $p_{\{a,b\}}=p_{\{a\}}=p_{\{b\}}=1$ and $p_\varnothing=0$, each single removal costs zero, but removing both costs one. Bounds therefore add over the successive contexts; in our runs no two withdrawals share a round (Figure~\ref{fig:objective}f), and re-testing withdrawn skills (Appendix~\ref{app:loop}) can restore a skill whose removal hurts. Admission, revision, and retrieval replacement enter through Eq.~\eqref{eq:app-round-identity}.

For one skill, let $a_k=\Pr_\rho(k\in K_x)$ and let $\mu_{\rho,k}$ be its mean current effect conditional on retrieval. Then $C_t=a_k\mu_{\rho,k}$. A ledger instead targets the tasks on which that skill was selected for testing. These distributions can differ even when the task curriculum is fixed, because skill selection depends on the task and ledger. They can also differ from the fixed evaluation distribution in Eq.~\eqref{eq:app-round-identity}.

The posterior bound transfers to the round cost up to a drift $d_k$: if a posterior concerns $\mu_k$, the desired current mean satisfies $\mu_{\rho,k}\le\mu_k+d_k$ under the same model, and withdrawal gives $\mathbb E[\mu_k^+]\le\epsilon$, then
\begin{equation}
 \mathbb E[C_t]\le a_k(\epsilon+d_k).
 \label{eq:app-cost-transfer}
\end{equation}
Matched policy, context, and task scope suffice for $d_k=0$. For effects in $[-1,1]$, a uniform effect drift $b_k$ and a task-distribution mismatch contribute at most $b_k+2\mathrm{TV}(\rho(\cdot\mid k\in K_x),\nu_k)$, where $\nu_k$ is the distribution of the tasks that test $k$. This follows by adding and subtracting the old effect under the new distribution and using the total-variation bound.

When several withdrawals are justified on successive contexts, let $\mu_j$ denote each retrieval-conditional mean effect and $a_j$ its evaluation retrieval mass. Under a common posterior model with $\mathbb E[\mu_j^+]\le\epsilon_j$ for each actual successive effect, linearity gives
$\mathbb E[C_t]\le\sum_j a_j\epsilon_j$.
Independence between the effects is unnecessary; the marginals only need to refer to the successive contexts.

The fixed evaluation distribution also matters. For changing distributions $\rho_t$, define inner gains and outer costs using $\rho_t$ within round $t$. Writing $J^{\nu}$ for evaluation under $\nu$, an additional term is
\begin{equation}
 D_t=J^{\rho_{t+1}}_{\mathcal S_{t+1}}(\theta_{t+1})-J^{\rho_t}_{\mathcal S_{t+1}}(\theta_{t+1}).
 \label{eq:app-task-drift}
\end{equation}
Adding and subtracting the second expectation at every transition yields a telescoping sum of $\Gamma_t-C_t+D_t$, so changing $\rho_t$ alters measured success without changing the Director. Keeping $\rho$ fixed in the deployed target removes $D_t$, while the Proposer still chooses training tasks adaptively.

\subsection{Tracking under separated scaffold changes}

\begin{lemma}[Two-time-scale tracking]\label{lem:twotime}
Let the inner state obey
\begin{equation}
 w_{t+1}=w_t+a_t\{h(w_t;\mathcal S_t)+M_{t+1}\}.
 \label{eq:app-tracking-recursion}
\end{equation}
Assume deterministic positive step sizes with $\sum_ta_t=\infty$ and $\sum_ta_t^2<\infty$; adapted iterates; predictable modes $\mathcal S_t$; and martingale-difference noise with uniformly bounded conditional second moment. Suppose all iterates lie in a compact set $D$, every mode has a locally Lipschitz mean field, and $D$ is forward invariant for its ODE. Assume finitely many modes, each with a Lyapunov-stable equilibrium $w^\star(\mathcal S)$ attracting every point of $D$. Finally, either mode changes stop, or their successive spacings on the clock $s_t=\sum_{u<t}a_u$ tend to infinity. For every $\zeta>0$, almost surely,
\begin{equation}
 \frac1{s_T}\sum_{t<T}a_t
 \mathbf 1\{\|w_t-w^\star(\mathcal S_t)\|>\zeta\}
 \longrightarrow0.
 \label{eq:app-occupation}
\end{equation}
\end{lemma}
\begin{proof}
The martingale $\sum_{t<n}a_tM_{t+1}$ has bounded total second moment and converges almost surely. Its increments over all sufficiently late finite clock windows therefore vanish uniformly. Boundedness of $h$ on $D$, $a_t\to0$, and the local Lipschitz property give the corresponding Euler approximation error. Applying the integral form of the ODE and Gronwall's inequality shows that, on any fixed-length interval without switching, the interpolation of the recursion uniformly approaches the ODE started at the same point. The constants can be chosen uniformly over the finite set of modes.

Fix $\zeta$. Stability provides a smaller neighborhood of each equilibrium whose ODE trajectories stay inside the $\zeta/2$ neighborhood. Attraction and compactness give a common finite time $H$ after which every trajectory starting in $D$ is inside an even smaller neighborhood. To see uniformity, choose an entry time for each initial point, use continuity of the finite-time flow to obtain a neighborhood with the same entry property, and take a finite subcover of $D$. Stability then preserves the entry property at later times. Taking the maximum over modes gives the common $H$.

On sufficiently late unswitched windows, the approximation error is smaller than both neighborhood margins. After the first window of length $H$, the stochastic interpolation is inside the inner neighborhood. Applying the same argument on consecutive windows keeps their endpoints inside it and their intervening paths within distance $\zeta$. Thus each switch contributes at most a fixed transient clock length, up to vanishing discretization errors. If $N(s)$ counts switches before clock time $s$, diverging spacings imply $N(s)/s\to0$: eventually every complete spacing exceeds any prescribed constant. The total transient fraction is consequently bounded by a vanishing initial fraction plus $O(HN(s)/s)$, proving Eq.~\eqref{eq:app-occupation}. If switching stops, the argument gives convergence to the final equilibrium.
\end{proof}

We hold the running statistics, such as the ledger and the reward normalization, fixed within each mode, so that each mode has an autonomous mean field. The finite-round accounting identity of Lemma~\ref{lem:decomp} holds without these conditions.

\subsection{Randomized contrasts, bootstrap means, and decisions}

\begin{lemma}[Randomization]\label{lem:random}
For a fixed task, policy, and candidate, assign exactly $n_v$ of $G$ rollout slots to the visible context uniformly at random, independently of their potential outcomes. Assume no interference between slots. Then
\begin{equation}
 \begin{aligned}
 \mathbb E[z_x\mid x,k]&=p^+-p^-=\Delta_k(x),\\
 \mathbb E[\bar Y\mid x,k]&=p^-+\tfrac{n_v}{G}\Delta_k(x).
 \end{aligned}
 \label{eq:app-randomization}
\end{equation}
For fixed slot-level potential outcomes, the design variance is
\begin{equation}
 \mathrm{Var}_{\mathrm{assign}}(z_x)
 =\frac{S_1^2}{n_v}+\frac{S_0^2}{n_h}-\frac{S_\Delta^2}{G},
 \label{eq:app-neyman-variance}
\end{equation}
where the finite-population variances use denominator $G-1$. Under independent superpopulation rollout sampling, the unconditional variance given the task is $V_+/n_v+V_-/n_h$.
\end{lemma}
\begin{proof}
Let $I_i$ mark a visible assignment and let $Y_i(1),Y_i(0)$ be fixed potential outcomes. Since $\mathbb EI_i=n_v/G$, the two observed part means have expectations equal to their corresponding population means. Their difference is the population mean treatment effect. Averaging over potential-outcome sampling yields the first identity, and the weighted average of the part means yields the second.

For the variance, $\mathrm{Var}(I_i)=n_vn_h/G^2$ and $\mathrm{Cov}(I_i,I_j)=-n_vn_h/[G^2(G-1)]$ for distinct slots. Write $z_x=\sum_i I_i c_i-\sum_iY_i(0)/n_h$ with $c_i=Y_i(1)/n_v+Y_i(0)/n_h$. Substitution of these covariances gives $n_vn_h\sum_i(c_i-\bar c)^2/[G(G-1)]$. Expanding this expression and using $S_\Delta^2=S_1^2+S_0^2-2S_{10}$ proves Eq.~\eqref{eq:app-neyman-variance}. Under independent rollout sampling, the two observed part means concern disjoint independent slots, so their variances add to $V_+/n_v+V_-/n_h$.
\end{proof}

A re-test group adds a withdrawn skill $k$ to the visible part, so by the same lemma its contrast estimates $p_{K_x\cup\{k\}}(x)-p_{K_x}(x)$. The design is blocked by task: the contrast removes the baseline term at a fixed task, while task selection can still change the average effect. A fixed mean group success does not identify $\Delta_k(x)$, since different $p^-$ values offset different effects in Eq.~\eqref{eq:app-randomization}.
For example, write $q_v=n_v/G$ and choose any $c\in(0,1)$. The probabilities $p^-=c-q_v d$ and $p^+=c+(1-q_v)d$ give mean group success $c$ and effect $d$. For all sufficiently small positive $d$, both are valid probabilities. Taking $d=0$ gives the same mean and zero effect. Thus the nonidentifiability is exact even under a fixed group ratio, while the randomized contrast distinguishes these cases in expectation.

\paragraph{What adaptive records estimate.}
Predictable task and skill selection does not destroy within-group randomization. To see what it preserves, index the successive records for a fixed skill version by $j$ and let $\mathcal H_{j-1}$ contain the information available before the next tested group is generated. Define $\mu_j=\mathbb E[z_j\mid\mathcal H_{j-1}]$, including the current task-selection law in that expectation. Then $e_j=z_j-\mu_j$ is a martingale difference, and for deterministic $n$,
\begin{equation}
 \mathbb E\left[\left(\bar z_n-\frac1n\sum_{j=1}^n\mu_j\right)^2\right]
 \le\frac1n.
 \label{eq:app-adaptive-ledger}
\end{equation}
For $i<j$, conditioning on $\mathcal H_{j-1}$ gives $\mathbb E[e_ie_j]=0$. Also $z_j\in[-1,1]$ implies $\mathbb E[e_j^2\mid\mathcal H_{j-1}]=\mathrm{Var}(z_j\mid\mathcal H_{j-1})\le1$. Expanding the square proves the bound, so the ledger mean tracks the average of the predictable effects $\mu_j$.

For a current target $\mu_{\mathrm{now}}$, the error is this sampling term plus the drift $n^{-1}\sum_j\mu_j-\mu_{\mathrm{now}}$; more records shrink the former, and the 30-cycle evidence window of Appendix~\ref{app:loop} keeps old records from dominating the latter. 

\paragraph{Bootstrap mean.}
Treat the ledger as an indexed collection of $n$ contrast records $z_1,\ldots,z_n$, allowing repeated task occurrences. For weights $w\sim\mathrm{Dirichlet}(1,\ldots,1)$, define $\Delta^{\mathrm{BB}}=\sum_jw_jz_j$. Conditional on the records,
\begin{equation}
 \begin{aligned}
 \mathbb E_{\mathrm{BB}}\Delta^{\mathrm{BB}}&=\bar z,\\
 \mathrm{Var}_{\mathrm{BB}}(\Delta^{\mathrm{BB}})
 &=\frac{\sum_j(z_j-\bar z)^2}{n(n+1)}.
 \end{aligned}
 \label{eq:app-bootstrap-moments}
\end{equation}
Indeed, $\mathbb Ew_j=1/n$, $\mathrm{Var}(w_j)=(n-1)/[n^2(n+1)]$, and distinct weights have covariance $-1/[n^2(n+1)]$; expanding the weighted sum proves both identities. Under a stationary exchangeable record model, this is the Bayesian-bootstrap posterior of the mean observed-contrast distribution. That mean equals the mean latent skill effect when randomization is unbiased on the same scope. No latent-effect likelihood is needed to define this mean.

For the adaptive ledger, we use the same Dirichlet construction as the posterior of the recorded comparison model, and the decisions below minimize posterior risk under it.

\begin{lemma}[Bayes decisions]\label{lem:bayes}
Fix a posterior $\Pi$ over a scope-specific mean effect $\Delta$. With losses $\Delta^+$ for withdrawal and $\epsilon$ for retention, withdrawal minimizes posterior expected loss exactly when $\mathbb E_\Pi[\Delta^+]\le\epsilon$, allowing withdrawal at a tie. With loss $1-\alpha$ for admitting a nonbeneficial skill and $\alpha$ for withholding a beneficial one, admission is optimal exactly when $\Pi(\Delta\le0)\le\alpha$, ties included.
\end{lemma}
\begin{proof}
The withdrawal and retention risks are precisely $\mathbb E_\Pi[\Delta^+]$ and $\epsilon$. For admission, put $q=\Pi(\Delta\le0)$. The two risks are $(1-\alpha)q$ and $\alpha(1-q)$; their comparison reduces to $q\le\alpha$. Finally $\mathbb E_\Pi\Delta\le\mathbb E_\Pi\Delta^+$, proving the signed-mean posterior bound used in Eq.~\eqref{eq:app-cost-transfer}.
\end{proof}

The positive part is taken \emph{after} forming the mean effect, and convexity places the resulting risk between two simpler quantities, $\bar z^+\le\mathbb E_{\mathrm{BB}}[(\Delta^{\mathrm{BB}})^+]\le n^{-1}\sum_jz_j^+$. The rule thus bounds the modeled aggregate harm over the test scope. Minimum record counts restrict eligibility, the withdrawal-first precedence resolves overlap between admission and withdrawal, and trial expiration (Appendix~\ref{app:loop}) is a separate management rule.

Versioning a skill by identifier, version, and content hash keeps evidence from pooling across edits to that skill, and the 30-cycle evidence window keeps it close to the current Director, context, and curriculum; the drift $d_k$ of Eq.~\eqref{eq:app-cost-transfer} carries the rest into the round cost. The prior pseudo-record enters only the exploration distribution $\Pi_k^0$ of Thompson selection, so decisions use the data-only $\Pi_k$, and the minimum-record gate keeps an untested skill from qualifying for a decision.

\subsection{The anchored curriculum and distinct pairs}

\begin{lemma}[Anchored curriculum]\label{lem:proposer}
Fix a finite candidate list, a full-support reference $\rho_0$, finite conditional expected rewards $\bar R_x$, and $\tau>0$. Over all distributions on the list, the unique maximizer of $\sum_x\rho_x\bar R_x-\tau\mathrm{KL}(\rho\|\rho_0)$ is the Gibbs distribution in Eq.~\eqref{eq:gibbs}. For two independent picks $a,b\sim\rho_\phi$, with conditional reward means depending only on the respective task,
\begin{equation}
 \mathbb E\left[\tfrac12(R_a-R_b)(s_a-s_b)\right]
 =\nabla_\phi\sum_x\rho_\phi(x)\bar R_x,
 \label{eq:app-iid-pairs}
\end{equation}
where $s_x=\nabla_\phi\log\rho_\phi(x)$ and rewards are held fixed when differentiating.
\end{lemma}
\begin{proof}
With $Z=\sum_x\rho_0(x)e^{\bar R_x/\tau}$, the objective is $\tau\log Z-\tau\mathrm{KL}(\rho\|\rho^\star)$, proving uniqueness. Expanding the pair estimator, the two self-terms each have expectation $\nabla_\phi\sum_x\rho_\phi(x)\bar R_x$. The cross-terms vanish by independence and $\mathbb E[s_x]=0$, and the factor one half gives Eq.~\eqref{eq:app-iid-pairs}.
\end{proof}

For the stated distinct-pair sampler, put $p_x=\rho_\phi(x)$ and assume at least two candidates with positive probability. Keeping the first pick and redrawing the second from the remaining candidates gives
\begin{equation}
 \Pr(a,b)=\frac{p_ap_b}{1-p_a},\qquad a\ne b.
 \label{eq:app-distinct-law}
\end{equation}
The unordered-pair probability sums both orders. Under the same task-local reward-mean assumption, the expectation of the raw estimator is
\begin{equation}
 \begin{aligned}
 &\sum_{a<b}p_ap_b\,w_{ab}(\bar R_a-\bar R_b)(s_a-s_b),\\
 &w_{ab}=\tfrac12\left(\frac1{1-p_a}+\frac1{1-p_b}\right).
 \end{aligned}
 \label{eq:app-distinct-gradient}
\end{equation}
This follows directly by multiplying the unordered-pair probability by the symmetric half-difference estimator. The independent-pick gradient is the same sum with $w_{ab}=1$. These pair-dependent factors cannot generally be absorbed into a single learning rate. For pair-dependent rewards, replace $\bar R_a-\bar R_b$ by $\mathbb E[R_a-R_b\mid\{a,b\}]$, with reward labels attached to tasks rather than pick order. 

A two-candidate example already separates the stationary points. Let $\bar R_1-\bar R_2=1$, $\rho_0=(1/2,1/2)$, $\tau=1$, and parameterize $p_1$ by its logit. The raw distinct-pair reward gradient is $1/2$, whereas the independent-pick gradient is $p_1(1-p_1)$. At the Gibbs logit of one, the KL gradient equals $p_1(1-p_1)$, so the raw distinct-pair regularized gradient remains nonzero. Gibbs optimality thus characterizes the ideal objective, while the stationary point of the practical update also depends on the pair weights, reward scaling, and trust region.

If both selected tasks supply training groups, the pooled task marginal within a list is
\begin{equation}
 \nu(x)=\tfrac12p_x\left(1+\sum_{a\ne x}\frac{p_a}{1-p_a}\right),
 \label{eq:app-pooled-task-law}
\end{equation}
obtained by averaging first- and second-position marginals. Conditioning further on selection of skill $k$ gives its test-task scope. Finally, a smaller expected reward implies a smaller Gibbs ratio $\rho^\star(x)/\rho_0(x)$, and the absolute probability also scales with the reference probability $\rho_0(x)$.

\subsection{The three propositions}
\label{app:props}

\begin{proof}[Proof of Proposition~1]
Lemma~\ref{lem:decomp} proves the accounting statement for every finite sequence. Appending a test-time step $\mathcal S_{T+1}=\varnothing$ with $\theta_{T+1}=\theta_T$ makes its left side the zero-skill success under $\rho$ for any run, with the benefit of the remaining scaffold as the last outer cost. Lemma~\ref{lem:twotime} proves the tracking statement under its assumptions, and Lemma~\ref{lem:mixed} identifies the context mixture as $q_x$, or $q'_x$ under separate last-skill centering.
\end{proof}

\begin{proof}[Proof of Proposition~2]
Apply Lemma~\ref{lem:tilt} to the policy on $K^-$ to obtain gain $\beta p^-(1-p^-)+O(\beta^2)$, or $\sqrt{2\delta p^-(1-p^-)}+O(\delta)$ at a common small budget, which grows with $\ell_{\mathrm{rec}}$ to first order; for a parametric Director, Eq.~\eqref{eq:app-tangent} bounds the accessible part. Lemma~\ref{lem:energy} identifies $4p^-(1-p^-)$ with normalized expected energy. Lemma~\ref{lem:equal} gives the shared-tilt optimum of the quadratic allocation problem. With stochastic Workers, Eq.~\eqref{eq:app-worker-noise} instead governs graph-controllable progress. The operational shrinkage, transfer, and collapse terms turn this target into the expected reward of Lemma~\ref{lem:proposer}.
\end{proof}

\begin{proof}[Proof of Proposition~3]
The first identity of Lemma~\ref{lem:random} proves conditional unbiasedness. Averaging it over the actual test-task distribution proves unbiasedness for that distribution's mean effect. The second identity mixes the post-withdrawal baseline with the skill effect, as in Eq.~\eqref{eq:confound}, while the contrast cancels the baseline within the task. Bootstrap aggregation and Lemma~\ref{lem:bayes} then give the decisions of Eq.~\eqref{eq:decide}, and Eq.~\eqref{eq:app-cost-transfer} carries them to the round costs.
\end{proof}

\section{Estimators and Implementation Details}
\label{app:impl}
\paragraph{Shrinkage and compute factor.}
The success rate $\tilde p_x$ in Eq.~\eqref{eq:rp} shrinks toward the running hidden-part success rate of the task's dataset with strength $m$, an exponential moving average over previous cycles with decay 0.9 that starts at one half. The positive term of Eq.~\eqref{eq:rp} is multiplied by $(\bar T_d/T_q)^{\eta}$, where $T_q$ is the mean duration of the task's rollouts and $\bar T_d$ the running mean duration of the dataset, so tasks are compared within a dataset by learning progress per unit of compute. A task without both parts, for instance one for which no skill is retrieved, is scored by $\ell$ of the shrunk success rate of the whole group, with the same collapse penalty.
\paragraph{Prior record and cost term.}
The Thompson draw of Eq.~\eqref{eq:post} adds one pseudo-record from a wide zero-mean prior to the bootstrap, so a skill with no data is drawn from the prior alone. The Director's reward subtracts $\lambda$ times the Workers' normalized cost from correctness; the contrasts $z_x$ always use raw correctness.
\paragraph{Director update.}
The Director's loss is the clipped surrogate of PPO on the advantages of Eq.~\eqref{eq:grpo}, whitened over the batch, which rescales every task by the same factor, and averaged over tokens, plus a KL anchor to the base model. Samples whose advantage is below a small threshold in magnitude, mostly from groups whose rollouts agree in correctness and differ only in cost, are dropped except for a small share kept for the anchor.
\paragraph{Proposer update.}
Each candidate list yields two distinct picks $a$ and $b$: draw $a$ from the choice policy and draw $b$ from that policy conditioned on $b\ne a$. The advantage of $a$ is half the reward difference $R_P(a)-R_P(b)$, scaled to unit root mean square over the batch, and $b$ takes its negative; the loss adds $\tau$ times the KL of the list's choice distribution to the reference. A trust region drops the policy term once the pair's tilt $(\log\rho_\phi(a)-\log\rho_\phi(b))-(\log\rho_0(a)-\log\rho_0(b))$ passes a bound in the direction the advantage pushes, so the choice within a list stays stochastic.
\paragraph{Graph canvas.}
Agents, layers, and relations are edited through the action set of Definition 2, and the canvas validates each edit before it is applied.

\section{Training Loop and Skill Ledger}
\label{app:loop}
Algorithm~\ref{alg:round} lists one round of \Rorch{}; every step cites the equation it implements.
An LLM editor revises skills, the least helpful first, and distills new ones from groups in which both parts fail, while withdrawals follow Eq.~\eqref{eq:decide}. A skill that the editor adds or revises enters the scaffold on trial under a new ledger key (identifier, version, and content hash), so its records start afresh; records for an unchanged skill identity remain pooled across Director updates. $\mathcal N_t$ admits it, a trial skill still unadmitted after $3n_{\min}$ records is discarded, and $\mathcal W_t$ is checked first, so the two rules never act on one skill in the same round.
A task with $K_x=\varnothing$ runs its $G$ rollouts without skills and adds no record. Each posterior uses the records of the last 30 cycles, and 30\% of the tasks re-test withdrawn skills, which $\mathcal N_t$ can restore.
\begin{algorithm*}[t]
\caption{One round of \Rorch{}}
\label{alg:round}
\small
\begin{algorithmic}[1]
\Require Director $\pi_\theta$, Proposer $\rho_\phi$, base models $\pi_{\mathrm{ref}},\rho_0$, scaffold $\mathcal S_t$ with summary $s_t$, ledger $\mathcal D_t$
\For{each candidate list $\mathcal C$ of the round}
  \State draw two distinct tasks from $\rho_\phi(\cdot\mid\mathcal C,s_t)$ \Comment{Appendix~\ref{app:impl}}
  \For{each selected task $x$}
  \State retrieve $K_x\subseteq\mathcal S_t$; on a re-test task, add one withdrawn skill to $K_x$ as $k_x$ \Comment{Appendix~\ref{app:loop}}
  \State otherwise, if $K_x\neq\varnothing$, draw $k_x$ by Thompson sampling; draw $\mathcal H_x$ uniformly, or set $\mathcal H_x=\varnothing$ if $K_x=\varnothing$ \Comment{Eq.~\eqref{eq:post}}
  \State set $K_i\gets K_x$ for all $i$; for $i\in\mathcal H_x$, set $K_i\gets K_x\setminus\{k_x\}$
  \State run the Director under $(x,K_i)$ and execute its graph on the Workers; record correctness $y_i$, $i\le G$
  \State compute $R_P(x)$ using Eq.~\eqref{eq:rp} or the skill-free fallback
  \If{$K_x\ne\varnothing$}
    \State $z_x\gets\bar y_{\mathcal V}-\bar y_{\mathcal H}$; append $(k_x,z_x)$ to $\mathcal D_t$ \Comment{Eq.~\eqref{eq:contrast}}
  \EndIf
  \EndFor
\EndFor
\State $\theta\gets$ clipped step on the advantages of Eq.~\eqref{eq:grpo} with the KL anchor to $\pi_{\mathrm{ref}}$
\State $\phi\gets$ distinct-pair step of Appendix~\ref{app:impl} with the KL anchor to $\rho_0$ and the trust region on the tilt
\For{each skill $k$ in $\mathcal S_t$ or under re-test with $n_k\ge n_{\min}$}
  \State $\Pi_k\gets$ Eq.~\eqref{eq:post}; withdraw $k$ if $k\in\mathcal W_t$, otherwise admit or restore it if $k\in\mathcal N_t$ \Comment{Eq.~\eqref{eq:decide}}
\EndFor
\State discard each trial skill still unadmitted after $3n_{\min}$ records
\State if a revision candidate with a defined posterior exists, revise the one with the lowest posterior benefit
\State distill new candidates from groups in which both parts fail
\State \Return $\theta,\phi,\mathcal S_{t+1},\mathcal D_{t+1}$
\end{algorithmic}
\end{algorithm*}

\begin{table*}[t]
\centering\small
\begin{tabular}{lll}
\toprule
Symbol & Meaning & Value\\
\midrule
$G$ ($n_h$, $n_v$) & rollouts per task (hidden, visible) & 5 (3, 2)\\
$\epsilon$ & withdrawal tolerance on the success rate & 0.02\\
$\alpha$ & admission risk & 0.05\\
$n_{\min}$ & minimum records per decision & 12\\
$B$ & Bayesian-bootstrap draws & 2000\\
-- & Thompson prior standard deviation & 0.25\\
-- & evidence window of the posterior (cycles) & 30\\
-- & share of tasks that re-test withdrawn skills & 0.3\\
$\gamma$ & discount when only the visible part succeeds & 0.5\\
$\lambda_P$ & penalty when all rollouts agree & 0.2\\
$m$ & shrinkage strength & 2\\
$\eta$ & compute-factor exponent & 0.5\\
$\lambda$ & Worker-cost weight in the Director reward & 0.005\\
$\tau$ & Proposer KL anchor & 1.0\\
-- & trust region on the Proposer's tilt (nats) & 4.0\\
-- & Director KL anchor; PPO clip (low, high) & 0.005; (0.2, 0.28)\\
-- & learning rate (Director, Proposer), cosine decay & $10^{-5}$, $10^{-5}$\\
-- & base model (Director, Proposer, Workers) & Qwen3.5-9B\\
-- & LoRA rank & 64\\
-- & tasks per cycle & 24 (2 lists of 8 per benchmark, 2 picks each)\\
-- & Director updates & 200\\
-- & sampling temperature (Director, Proposer, Worker) & 1.0, 1.0, 0.3\\
\bottomrule
\end{tabular}
\caption{Hyperparameters of \Rorch{}.}
\label{tab:hyper}
\end{table*}
\section{Experimental Details}
\label{app:exp}
\paragraph{Benchmarks and splits.}
The six IID benchmarks, HotpotQA \citep{yang2018hotpotqa}, NQ-Open \citep{kwiatkowski2019natural,lee2019latent}, MedQA \citep{jin2021disease}, AIME, MBPP+ \citep{austin2021program,liu2023your}, and ALFWorld \citep{shridhar2020alfworld}, supply the training tasks and the IID tests; AIME trains on the pooled problems of earlier years and is tested on AIME 2026. The six OOD benchmarks, TriviaQA \citep{joshi2017triviaqa}, MuSiQue \citep{trivedi2022musique}, GPQA \citep{rein2023gpqa}, MATH-Hard \citep{hendrycks2021measuring}, SWE-Bench \citep{jimenez2024swe}, and WebShop \citep{yao2022webshop}, are used only for testing. Each test set has 128 items (AIME 2026: 30) disjoint from the training tasks, so one run's 0/1 metric lies on a $k/128$ grid ($k/30$ for AIME 2026) and a five-run mean on a $k/640$ grid ($k/150$). The training pools and sizes are released with the code. The training curves in Figure~\ref{fig:transfer}a use a separate held-out set of 32 tasks from each IID dataset (AIME: 30 problems of earlier years), disjoint from the training tasks and distinct from the IID and OOD test sets.
\paragraph{Metrics.}
HotpotQA, NQ-Open, TriviaQA, and MuSiQue report exact match (EM) and F1 of the final answer, with Ans EM in Table~\ref{tab:ablate}; MedQA, AIME 2026, GPQA, and MATH-Hard report accuracy; MBPP+ reports Pass@1; ALFWorld and WebShop report the success rate (SR); and SWE-Bench reports the share of resolved issues.
\paragraph{Baselines and fairness.}
Qwen3.5-9B \citep{qwen3.5} and DeepSeek-V4-Flash are prompted directly. SFT and GRPO \citep{shao2024deepseekmath}, the latter with dynamic sampling, train Qwen3.5-9B on the same training tasks, and each keeps the configuration whose trained weights score best under its default recipe. The other baselines are workflow search (AFlow \citep{zhang2025aflow}), RL-trained orchestration (AgentFlow, FlowSteer \citep{li2026flow,zhang2026flowsteer}), skill internalization (SKILL0, PATS \citep{lu2026skill0,shi2026pats}), and self-play with skill evolution (SESA, MetaSkill \citep{fu2026self,wang2026metaskill}), and each runs the best configuration reported in its paper. All methods share the frozen Workers, training tasks, and compute budget; \Rorch{} is tested at zero skills.
\paragraph{Ablation variants.}
Each ablation changes one component and keeps the rest of \Rorch{} fixed, including the Workers, the training tasks, the training rollouts, and the zero-skill test. In each group of Table~\ref{tab:ablate}, the first row removes the whole component: GRPO only for Recursive Self-Play Orchestration, a failure reward for Recursive Learnability, and a fixed scaffold for Skill Withdrawal. The other rows remove one part of the orchestration (the skills or the Proposer's picks), replace only the Proposer's reward, or replace only the withdrawal rule, whose thresholds are tuned on validation.
\paragraph{Proposer objectives.}
Figure~\ref{fig:objective}a--e replace only the Proposer's objective, by uniform sampling, dynamic sampling (DAPO \citep{yu2026dapo}), selective rollout (GRESO \citep{zheng2026act}), learnability on the current scaffold (SFL \citep{rutherford2024no}), a solve-rate reward (Absolute Zero, AZR \citep{zhao2026absolute}), or an uncertainty reward (R-Zero \citep{huang2026r}), and keep the Director update, the scaffold, the withdrawal rule, and the training rollouts fixed; the Uniform, DAPO, SFL, and \Rorch{} rows of Figure~\ref{fig:objective}a repeat Table~\ref{tab:ablate}. Every objective trains 24 groups of five rollouts per cycle, DAPO's discarded rollouts count toward its cost, and scores are five-run means at zero skills.
\paragraph{Backbone transfer.}
For Figure~\ref{fig:transfer}e, the Director keeps the weights trained with Qwen3.5-9B Workers, and only the Workers are replaced by each of six other backbones, with skills off. Frozen prompts the backbone directly, and +\Rorch{} runs the same backbone under the trained Director.
\paragraph{Training dynamics.}
Figure~\ref{fig:transfer}a evaluates the checkpoint of every training cycle on the held-out set, once with the current skills and once with none. Figure~\ref{fig:transfer}b counts the distinct graph types the Director builds over the six IID datasets. A candidate task is unseen if it has never been attempted; otherwise its previous success rate $p$ over its earlier rollouts makes it unsolved ($p=0$), learnable ($0<p<1$), or solved ($p=1$). Figure~\ref{fig:transfer}c sums the Proposer's probability over the candidates of each class, and Figure~\ref{fig:transfer}d gives its probability on solved candidates as a moving average over 10 cycles, beside the untrained model with the same prompt and random choice (the share of solved candidates).
\paragraph{Selection evidence and withdrawals.}
Each candidate list of eight tasks yields two distinct picks (Appendix~\ref{app:impl}). Figure~\ref{fig:objective}e counts the 2,220 picks of seen tasks over cycles 50--199 by their previous success rate, and a new group is mixed if its rollouts contain at least one success and one failure. These counts include both picks of each list, the second drawn from the remaining candidates (Eq.~\eqref{eq:app-distinct-law}), whereas Figure~\ref{fig:transfer}d gives the policy's probability on solved candidates as a moving average over 10 cycles. Figure~\ref{fig:objective}f measures the held-out gain of a dataset's three original seed cards, used together, over no cards, once at checkpoint 0 and as the mean over the 10 checkpoints around each withdrawal with its 95\% interval. Each row names the withdrawal cycle and the card; \emph{restored} marks the MedQA topology card, which re-testing restored after its withdrawal at cycle 6. This is dataset-level evidence; each withdrawal follows the posterior of the withdrawn card's contrast (Eq.~\eqref{eq:decide}).
\paragraph{Hyperparameters.}
Table~\ref{tab:hyper} lists every constant that appears as a symbol in the main text; the PPO clip and the learning rate set the step size $\beta$.

\paragraph{Compute.}
\Rorch{} trains on A100 GPUs for 200 cycles, about 106 A100 hours measured in training, twice its 100-cycle cost. Figure~\ref{fig:objective}c,d report the selection cost of the Proposer objectives: extra and total rollouts per cycle, rollouts and tokens per effective group, and A100 hours per 100 cycles. An effective group mixes successes and failures, so its GRPO advantages are not all zero.
\paragraph{Repeated runs.}
Table~\ref{tab:main} reports the mean and standard deviation over five runs, and Table~\ref{tab:ablate} and Figures~\ref{fig:transfer}e and~\ref{fig:objective}a report five-run means.

\end{document}